\documentclass[10pt,journal,twocolumn]{IEEEtran}
\ifCLASSOPTIONcompsoc
  \usepackage[nocompress]{cite}
\else
  \usepackage{cite}
\fi
\ifCLASSINFOpdf
\else
\fi

\usepackage{bbm}
\usepackage{epsfig}
\usepackage{graphicx}
\usepackage{amsmath,amssymb}
\usepackage{eucal,bibspacing}
\usepackage{cite}
\usepackage{cs}
\usepackage{mathsymb}
\usepackage{colortbl}
\usepackage{color}
\usepackage{xcolor}
\usepackage{algorithmic}

\usepackage{multirow}
\usepackage{pbox}
\usepackage{wrapfig}
\usepackage{caption}
\usepackage[linesnumbered,algoruled,boxed,lined]{algorithm2e}

\usepackage{algorithm}
\def\bf{{\boldsymbol f}}

\begin{document}

\title{Jacobian Norm with Selective Input Gradient Regularization for Improved and Interpretable Adversarial Defense}

\author{Deyin Liu, Lin Wu, Haifeng Zhao, Farid Boussaid, Mohammed Bennamoun, Xianghua Xie

\IEEEcompsocitemizethanks{\IEEEcompsocthanksitem 
\protect\\ D. Liu and H. Zhao are with Anhui Provincial Key Laboratory of Multimodal Cognitive Computation, School of Artificial Intelligence, Anhui University, Hefei 230039, China. E-mail: deyinde@126.com; senith@163.com.
\protect\\ F. Boussaid and M. Bennamoun are with The University of Western Australia, Perth 6009, Australia. E-mail: \{Farid.Boussaid; Mohammed.Bennamoun\}@uwa.edu.au.
\protect\\ L. Wu and X. Xie are with School of Computer Science, Swansea University, SA1 8EN, UK. E-mail: \{l.y.wu;x.xie\}@swansea.ac.uk.
%\IEEEcompsocthanksitem J. Doe and J. Doe are with Anonymous University.
}% <-this % stops an unwanted space
%\thanks{Manuscript received April 19, 2005; revised September 17, 2014.}
}

\IEEEtitleabstractindextext{%
\begin{abstract}
Deep neural networks (DNNs) are known to be vulnerable to adversarial examples that are crafted with imperceptible perturbations, i.e., a small change in an input image can induce a mis-classification, and thus threatens the reliability of deep learning based deployment systems. Adversarial training (AT) is often adopted to improve robustness through training a mixture of corrupted and clean data. However, most of AT based methods are ineffective in dealing with \textit{transferred adversarial examples} which are generated to fool a wide spectrum of defense models, and thus cannot satisfy the generalization requirement raised in real-world scenarios. Moreover, adversarially training a defense model in general cannot produce interpretable predictions towards the inputs with perturbations, whilst a highly interpretable robust model is required by different domain experts to understand the behaviour of a DNN. In this work, we propose a novel approach based on Jacobian norm and Selective Input Gradient Regularization (J-SIGR), which suggests the linearized robustness through Jacobian normalization and also regularizes the perturbation-based saliency maps to imitate the model's interpretable predictions. As such, we achieve both the improved defense and high interpretability of DNNs. Finally, we evaluate our method across different architectures against powerful adversarial attacks. Experiments demonstrate that the proposed J-SIGR confers improved robustness against transferred adversarial attacks, and we also show that the predictions from the neural network are easy to interpret.
\end{abstract}

}

\maketitle

\IEEEdisplaynontitleabstractindextext

\IEEEpeerreviewmaketitle

\section{Introduction}\label{sec:intro}

\IEEEPARstart{R}{ecent} years have seen deep neural networks (DNNs) achieve impressive performance in a myriad of image recognition tasks, e.g., image classification on real-world benchmarks. However, DNNs could be easily fooled by crafted imperceptible perturbations aimed at driving the network to make a wrong prediction. Such a vulnerability poses an obstacle in the deployment of deep learning systems for real-world applications given the security-sensitivity and legal ramifications \cite{Crime-prediction}. Since this vulnerability was first identified by Szegedy et al. \cite{FGSM}, many techniques for generating a group of malicious examples have been proposed including i) \textit{white-box attacks} (e.g., Projected Gradient Descent (PGD) \cite{PGD}, Fast Gradient Sign Method (FGSM) \cite{FGSM} and C\&W \cite{C-W-attack}) which require knowledge of the deep learning parameters; and ii) \textit{black-box attacks} \cite{Substitute,ZOO,Orthogonal-bbx} can attack the DNNs even without any knowledge of the model parameters. This is due to the \textit{transferability} of adversarial examples: examples generated to fool one model tend to fool \textit{all} models trained on the same dataset \cite{Space-TAE,Attack-Fool}.

Currently, the best defense models available to counter adversarial attacks are based on \textit{adversarial training} (AT), which trains the 
models on a mixture of clean and adversarial examples to better classify potentially adversarial examples during test time \cite{Adversarial-features}. Although a number of effective defenses employing AT have been proposed \cite{Feature-scatter,AT-free}, they rely on a computationally expensive brute force solution to generate the potent adversarial examples \cite{Logit-pairing}. Besides, a study \cite{Ensemble-AT} shows that the robustness provided by AT can be circumvented by randomizing or transferring perturbations from other models, albeit ensemble helps. This has raised concerns about the generalization ability of adversarially trained models under the attacking of adversarial examples simply modified from a unknown distribution. For example, a robust DNN model adversarially trained on gradient-generated examples should still remain robust when these examples are further crafted by Gaussian randomness on the feature dimensions (The dversarially generated example with a new additive Gaussian noise can be viewed as a transferred one from its source \cite{PNI}).

Besides robustness, domain experts are also concerned with the interpretability of a DNN's predictions. Interpretability is of particular importance in domains with safety requirements which need to know how a model is trained and used. For example, medical specialists are intriguing to know how the model responds to training data from different hospitals. To be more interpretable, some methods proposed to regularize the input gradients \cite{Regularization-robustness,Input-Grad-R}, which can illuminate the regions of confidence predicted by the DNNs. For example, \cite{Input-Grad-R} has shown that by training a model to have smooth input gradients with fewer extreme values (including the background pixels), the prediction of a DNN will be more interpretable. However, this endeavor \cite{Input-Grad-R} smoothed out all the gradients without showing the most appropriate interpretability of DNNs, i.e., such method \cite{Input-Grad-R} is limited in presenting the network's response to small input variations. Other techniques such as integrated gradients \cite{IntegGrad} and SmoothGrad \cite{SmoothGrad} can also generate smoother and more interpretable prediction confidence but they do not capture the response of a deep neural network under input variations. As a matter of fact, the local behavior can simulate how the network responds to small perturbations of its inputs.

In this paper, we aim to improve both robustness and prediction interpretability in the presence of small input perturbations. This is achieved by simultaneously minimizing the Jacobian norm of the whole network as well as regularizing the input gradients. Specifically, we introduce the notion of linearized robustness, which sets the bounds on the network response towards a perturbed example. And we show that the robustness can be approximated by the Jacobian norm, which is computed as the gradients of the network prediction logits with respect to the input. However, directly using the raw input gradients could be noisy and hard to interpret. Inspired by \cite{Regularization-robustness} which has shown that gradient suppression together with selected features can explain the robustness of a model, we propose to regularize the input gradients to provide a better input space. Then, we train the network using the Jacobian's Frobenius norm, which sets the bounds on the response of the network layers to input perturbations. We empirically show how the Jacobian norm relates to the linearized robustness of the DNNs, and how to leverage it into the adversarial training to mitigate the adversarial effects.

To further make the prediction more interpretable, we propose to use the perterbation-based saliency map, which is taken as the salient Jacobian matrix that can relate the gradients to the corrupted input image and semantically visualize the discriminative portions of that image \cite{Robustness-odds}. These Jacobians, computed during training for different robust models, can be regarded as visual resemblers to the corresponding images, and a previous study has shown that the saliency map is a result of robustness \cite{Robust-saliency}. This strategy is demonstrated to be robustness across different transferred adversarial examples. The contributions of this paper can be summarized as:
\begin{itemize}
\item We propose a novel approach to achieve both improved robustness and high interpretability of DNNs under adversarial attacks. The proposed approach effectively leverages the Jacobian norm and selective input gradient regularization, which explicitly describes the network responses to input perturbations. 
\item We investigate the relationship between a Jacobian norm  and the linearized robustness \footnote{(The distance from the input image to the decision boundary)} of DNNs. Based on perturbation-based saliency maps, our results give insights into the prediction confidence of DNNs, which relates to the adversarial effects.
\item Our method improves the robustness of DNNs towards transferred adversarial examples across multiple architectures and different attacks.
\end{itemize}

The rest of this paper is organized as follows. Section \ref{sec:related} reviews the recent related works. Section \ref{sec:method} details the proposed method, and Section \ref{sec:experiment} presents extensive experiments to evaluate our method. Finally, we conclude the paper in Section \ref{sec:con}.

\section{Related work}\label{sec:related}

\subsection{Adversarial training based methods}
There is a sizable body of work proposing various attack and defense mechanisms for the adversarial setting. Among them, the current unbroken defenses are based on adversarial training (AT) \cite{FGSM,TRADES,PGD}, which uses adversarial examples as training data to protect DNNs against a range of adversarial attacks. For example, projected gradient descent (PGD) is one such strong defense that is able to generate universal adversarial examples using a first-order approach \cite{PGD}. A more recent work \cite{TRADES} encourages the decision boundary to be smooth by adding a regularization term to reduce the difference between the predictions on natural and adversarial examples. Qin et al. \cite{Adv-local-linearization} smoothened the loss landscape through local linearization by minimizing the prediction difference between the clean and adversarial examples. While the various aforementioned approaches can improve the AT, they require the generation of sufficient adversarial examples for training. This results in a prohibitive computational cost, which is proportional to the number of steps needed to generate the adversarial examples. In addition, it requires a back-propagation for each iteration. To strengthen DNNs under adversarial attacks, a biologically-inspired approach \cite{Biological-adversarial} was introduced to learn flat, compressed representations that are sensitive to a minimal number of input dimensions. Unlike \cite{Biological-adversarial}, this paper introduces a simpler yet effective approach for model regularization that is based on input gradient regularization. A concurrent method is \cite{Input-gradients-robustness}, which can improve robustness by imposing the input gradient regularization. However, performing such gradient with respect to a high dimensional input from back-propagation is quite time-consuming. In contrast, the proposed method approximates the linearized robustness of neural networks via the penalization of a classifier's Jacobian norm. Such a Jacobian norm derives salient gradient maps to selectively activate the most discriminative gradients.

\subsection{Regularization for robustness}
To defend against adversarial examples, provable defenses promote the concept of improving model robustness through regularization. A well-known strategy includes noise injection, which is a variant of dropout weights \cite{Dropconnect} or activations \cite{Dropout}. Several works have investigated the benefits of using a regularization term on top of the standard training objective to reduce the Jacobian's Frobenius norm. Such a term aims to reduce the adversarial effect on the model predictions caused by input perturbation. For instance, Hoffman et al. \cite{Robust-jacobian} proposed an efficient method to approximate the input-class probability through the output Jacobians of a classifier so as to minimize the computational cost associated to these Jacobian norms. Tsipras et al. \cite{Robustness-odds} observed that adversarially trained models produce salient Jacobian matrices that loosely resemble the input images whilst less robust models have noisier Jacobians. Etmann et al. \cite{Robust-saliency} interpreted linearized robustness as the alignment between the Jacobian and the input image, and trained a robust model by using double back-propagation. In comparison to these methods \cite{Robustness-odds,Robust-saliency}, our work offers the following merits: 1) it focuses on the local linearized robustness of neural networks to provide an interpretation of the network's response to the inputs via Jacobian norm; 2) the proposed selective input gradient regularization explicitly measures the degree of robustness, which improves the interpretability of the network prediction; and 3) our method is computationally more efficient in calculating the input gradients and also highly interpretable to understand the network's prediction. 

\section{Preliminary}\label{sec:pre}
In this section, we formalize the notations, definitions used in this paper. We also briefly present the baseline attacks and defenses against which we will evaluate our method. 

\subsection{Definition}
Let a classification model $F_{\theta}(\mathbf{X}): \mathbf{X} \mapsto \hat{\mathbf{Y}} \in \mathbb{R}^{N\times K}$ map the inputs $\mathbf{X}\in \mathbb{R}^{N\times D}$ to the output probabilities for $K$ classes, where $\theta$ represents the classifier's parameters and $\hat{\mathbf{Y}} \in \mathbb{R}^{N\times K}$ returns the predictions of $F_{\theta}$. To train the model $F_{\theta}$, we aim to find a set of parameter $\theta^*$ that minimizes the total distance between the predictions $\hat{\mathbf{Y}}$ and the one-hot encoded true labels $\mathbf{Y}\in \mathbb{R}^{N\times K}$ on a training set:
\begin{equation}
    \theta^* = \arg\min_{\theta} \sum_{n=1}^N \sum_{k=1}^K -\mathbf{Y}_{n,k} \log F_{\theta}(\mathbf{X}_{n,k}),
\end{equation}
which can also be written as $\arg\min_{\theta} H(\mathbf{Y},\hat{\mathbf{Y}})$, where $H$ is the sum of the cross entropies between the predictions and the true labels.

Given the input $\mathbf{x} \in \mathbb{R}^{h\times w\times c}$ of a DNN, one can define the Jacobian matrix $J$ with respect to $\mathbf{x}$ as
\begin{equation}
J(\mathbf{x}):=\nabla_{\mathbf{x}} F_{\theta}(\mathbf{x})=\left[ \frac{\partial F_{\theta}(\mathbf{x})}{\partial \mathbf{x}_1},\ldots, \frac{\partial F_{\theta}(\mathbf{x})}{\partial \mathbf{x}_D} \right],
\end{equation}
where $D=h\times w \times c$ is the dimensionality of $\mathbf{x}$. While the DNNs can be trained empirically to perform well on the training data, their accuracy degrades sharply in the presence of adversarial examples. Given a small perturbation $\mathbf{z}$ applied to the input $\mathbf{x}$, a model is still deemed robust against this attack if it satisfies
\begin{equation} 
\begin{aligned}
& \arg\max_{k\in K} F^k_{\theta} (\mathbf{x})=\arg\max_{k\in K} F^k_{\theta}(\mathbf{x} +\epsilon \mathbf{z}), \\
&\forall \epsilon \in B_p(\varepsilon)=\epsilon: ||\epsilon ||_p \leq \varepsilon,
\end{aligned}
\end{equation} 
where $\varepsilon$ is the scaling factor, and $p=\infty$. To improve the model's robustness, adversarial training \cite{FGSM} tries to find a distribution match between the training data and the adversarial test data. Specifically, adversarial training attempts to minimize the loss function as: 
\begin{equation}\label{eq:AT-loss}\small
\min_{\theta}\rho(\theta), \mbox{where} ~~ \rho(\theta)=\mathbb{E}_{(\mathbf{x},\mathbf{y})} \left[ \max_{\epsilon \in B(\varepsilon)} H ( F_{\theta}(\mathbf{x}+\epsilon \mathbf{z}), \mathbf{y}) \right],
\end{equation}
where the inner maximization terms are usually obtained by performing an iterative gradient-based optimization, such as the projected gradient descent (PGD) \cite{PGD}.

\subsection{Attacks}
We consider two widely adopted gradient-based attacks and one Jacobian-based attack.

\paragraph{Fast Gradient Sign Method (FGSM) \cite{FGSM}} This method can generate adversarial examples by perturbing the inputs s increase the local linear approximation of the loss function: $\mathbf{x}_{FGSM}= \mathbf{x}+ \epsilon\cdot\mbox{sign} \nabla_{\mathbf{x}} H(\mathbf{y},\hat{\mathbf{y}})$. If $\epsilon$ is small (e.g., $\epsilon=0.01$), adversarial examples become indistinguishable to a human, but a neural network performs significantly poor with them. To perform this attack, one can iteratively use a small $\epsilon$ to induce misclassifications by following the non-linear loss function in a series of small linear steps \cite{Adv-examples-physical}.

\paragraph{Projected Gradient Descent (PGD) \cite{PGD}} The PGD provides a white-box and yet stronger attack than the previous iterative based method (i.e., FGSM \cite{FGSM}). PGD firstly uniforms the random perturbation as the initialization. Then, an adversarial example is found after several iterations by taking the form as $\mathbf{x}_{PGD}^{t+1}$=$\prod_{\mathbf{x}+S} [\mathbf{x}_{PGD}^t + \epsilon\cdot\mbox{sign} \nabla_{\mathbf{x}} H(\mathbf{y},\hat{\mathbf{y}})]$,
where $\prod$ is the projection operator that clips the input at positions around the predefined perturbation range. $\mathbf{x}+S$ represents the perturbation set, and $\epsilon$ is the gradient step.

\paragraph{Jacobian-based Saliency Map Attack (JSMA) \cite{JSMA}} The JSMA iteratively searches for pixels of the input to change such that the probability of the target label is increased and the probability of all other labels are decreased. Such a method can produce examples that have only been modified in a fraction of feature dimensions, which are hard for humans to detect.

\subsection{Defenses}

We consider two baseline defenses: adversarial training and defensive distillation. Defensive methods that are not architecture-agnostic \cite{Feature-squeezing} or simply rejecting adversarial examples \cite{Biological-adversarial} are not considered in this paper.

\paragraph{Adversarial training \cite{Adversarial-features}} Adversarial training (AT) can enhance robustness by injecting adversarial examples into the training process. We follow the implementation in \cite{Cleverhans}, where we augment the network to run the FGSM \cite{FGSM} on the training batches, and compute the average loss on both the normal and adversarial examples as the loss function of the model. To inhibit the FGSM attack \cite{FGSM}, gradients are not allowed to propagate and FGSM perturbations are computed with respect to the predicted labels (instead of the true labels) to prevent label leaking. 

\paragraph{Defensive distillation \cite{Adversarial-features}} Distillation can be used as a defense technique by first using the one-hot ground truth labels to train an initial model and subsequently utilizing the initial model's softmax probability outputs. Since distillation extracts class knowledge from these probability vectors, this knowledge can be transferred into a different DNN architecture by annotating the inputs in the training dataset of the second DNN using classification predictions from the first DNN. This idea is formulated to improve the resilience of DNNs in the presence of perturbations \cite{Distillation-Defense}. Within a softmax layer, we divide all of the logit network output (which we call $\hat{z}_k$) by a temperature $T$: $F_{T,\theta}(\mathbf{X}_{n,k})=\frac{e^{\hat{z}_k(\mathbf{X}_{n,k})/T}}{\sum_{i=1}^K e^{\hat{z}_i(\mathbf{X}_{n,k})/T}}$, where $F_{T,\theta}$ denotes a network output in the form of a softmax vector with temperature $T$. The predictions will converge to $1/K$ as $T\rightarrow \infty$. Thus, the distillation based defense can be formulated as 
\begin{equation}
\begin{aligned}
   & \theta^0 = \arg\min_{\theta} \sum_{n=1}^N \sum_{k=1}^K -\mathbf{Y}_{n,k} \log F_{T,\theta}(\mathbf{X}_{n,k}),\\
    & \theta^* = \arg\min_{\theta} \sum_{n=1}^N \sum_{k=1}^K -F_{T, \theta^0}(\mathbf{X}_{n,k}) \log F_{T,\theta}(\mathbf{X}_{n,k}).
\end{aligned}
\end{equation}

\section{Proposed approach}\label{sec:method}

\begin{figure*}[t]
  \centering
  \includegraphics[width=0.8\linewidth]{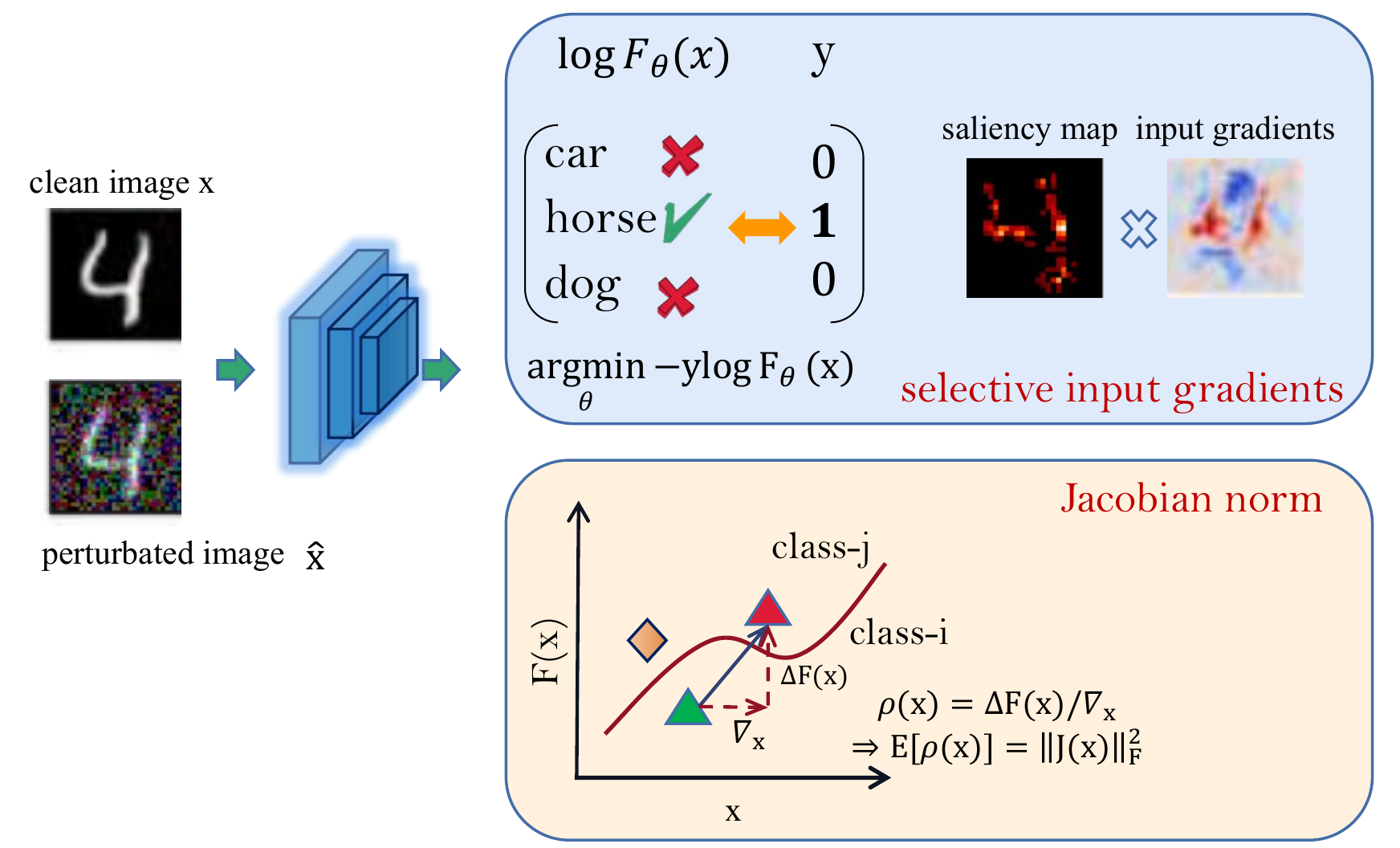}
  \caption{The proposed scheme for adversarial robustness based on Jacobian normalization and selective input gradient regularization (J-SIGR). The Jacobian normalization sets the linear robustness bounds for perturbations. The selective input gradient regularization is based on perturbation-based saliency map to not only encourage the insensitivity of the input space but also improves the interpretability.}\label{fig:framework}
\end{figure*}

In this section, we detail the proposed method (Fig. \ref{fig:framework}), which leverages the Jacobian norm with selective input gradient regularization (namely J-SIGR) for both improved defense and interpretability of a deep neural network under powerful attacks.

\subsection{Jacobian norm}

In the following, we will study the relationship between the Jacobian norm (JN) based regularization term and the notion of linearized robustness. Since adversarial perturbations are small variations that change the predicted result of a neural network classifier, it is sensible to define the robustness towards adversarial perturbations via the distance of the clean image to the nearest perturbed image which may cause the incorrect classification. When such distance gets smaller, the perturbed and its clean counterpart are more indistinguishable for a neural network, and thus the prediction of the neural network will be correct.

\textbf{Linearized adversarial robustness bound:}
Let $i^*=\arg\max_i F^i_{\theta} (\mathbf{x})$ and $j^*=\arg\max_{j\neq i^*} F^j_{\theta} (\mathbf{x} + \epsilon \mathbf{z})$ be the top prediction of the input $\mathbf{x}$ and its corrupted sample $\hat{\mathbf{x}}=\mathbf{x} + \epsilon \mathbf{z}$, respectively. Here, $\hat{\mathbf{x}}$ is formed by small additive perturbations with Gaussian distribution $\mathbf{z}\sim \mathcal{N}(0,\sigma^2)$. The linearized adversarial robustness is upper-bounded by the Jacobian norm $||J(\mathbf{x})||_F^2$ with respect to $\mathbf{x}$.

\textit{Proof.} Denoting $F^i_{\theta}(\mathbf{x})$ as the logits value of class $i$ in a classifier $F$ \footnote{For notation simplicity, we omit $\theta$ in the following.} for $\mathbf{x}$ with $\epsilon \ll 1$, then its linearized robustness can be expressed as 
\begin{equation}
\rho(\mathbf{x}):=\min_{j^*\neq i^*} \frac{ F^{i*}(\mathbf{x} )-F^{j*}(\mathbf{x}) }{||\nabla_{\mathbf{x}} F^{i*}(\mathbf{x} )-\nabla_{\mathbf{x}} F^{j*}(\mathbf{x}) || }. 
\end{equation}
Denoting $g:=\nabla_{\mathbf{x}} (F^{i*}- F^{j*})(\mathbf{x})$ as the Jacobian w.r.t the difference of two logits and $\alpha(\mathbf{x},g)=<\mathbf{x},g>$ as the alignment between the Jacobian and the input, then we have $\rho(\mathbf{x})\leq \frac{\alpha(\mathbf{x},g) +C }{||g||}$, where $C$ is a positive constant. Therefore, $\rho(\mathbf{x})\leq \frac{J(\mathbf{x})+ g +C}{|| g ||}$, where $J(\mathbf{x})$ is the Jacobian of the network output w.r.t the input $\mathbf{x}$. 

We can now treat the term $\hat{\rho}= \mathbf{z}^T J(\mathbf{x})^T J(\mathbf{x}) \mathbf{z}$ as one sample stochastic trace estimator for $Tr(J(\mathbf{x})^T J(\mathbf{x}))$ with a Gaussian variable $\mathbf{z}$:
\begin{equation}
\mathbb{E}_{\mathbf{z}} [\rho]=\frac{Tr(J(\mathbf{x})^T J(\mathbf{x})\mathbb{E}[\mathbf{z} \mathbf{z}^T])}{ ||g||}=\frac{|| J(\mathbf{x}) ||_F^2}{|| g ||}. 
\end{equation}
Taking expectation over $m$ samples of a mini-batch $X$, we have $\mathbb{E} [\rho]=\mathbb{E}_{\mathbf{x}} [||J_X ||]_F^2$, where $||\cdot ||_F^2$ represents the Frobenius norm. We remark that the above assumption holds true given that the neural network can be locally approximated by a linear model \cite{Robust-saliency}. For classifiers built on locally affine score functions, as in the case of most neural networks using ReLU or leaky ReLU activations, the decision boundary can be computed, provided that the locally affine region around the point $\mathbf{x}$ is sufficiently large. As proved in \cite{Robust-saliency}, for a classifier defined with a locally affine score function, the decision boundary between the clean and the perturbed data is close in the Euclidean space when their respective input signals are also close enough in the Euclidean space \cite{Lya-Theory}. Thus, the linearized robustness holds approximately as long as the linear approximation to the network's score function is sufficiently plausible in the relevant neighborhood of $\mathbf{x}$.

\begin{algorithm}[tb]
\caption{The proposed adversarial robust model based on Jacobian Norm and Selective Input Gradient Regularization (J-SIGR).}
\label{alg:algorithm-J-SIGR}
\textbf{Inputs}: Training set $D_{train}=\{\mathbf{x}_i,\mathbf{y}_i\}^N_{i=1}$. \\
\textbf{Parameters}: $\theta$, $\lambda_j$, $\lambda_m$ $\epsilon$.\\
\textbf{Outputs}: Classification model $F_{\theta}(\cdot)$ parameterized by $\theta$.\\
/*Initialization*/
\begin{algorithmic}[1] %[1] enables line numbers
\STATE Initialize $\theta$ by using a pre-trained network architecture; 
\STATE Set $\lambda_j=\lambda_m=0.5;\epsilon=0.3$.
\WHILE{each iteration or a condition is met}
\STATE Sample $(\mathbf{x},\mathbf{y}) \sim D_{train}$; /*Input to the DNN*/
\STATE Generate the noise perturbation $\mathbf{z}\sim \mathcal{N}(0,\sigma^2)$;
\STATE $\hat{\mathbf{x}}=\mathbf{x}+\epsilon \mathbf{z}$; /*Generate a perturbed sample*/
\STATE $\nabla_{\mathbf{x}} F_{\theta}(\hat{\mathbf{x}})$; /*Compute the perturbation-based saliency map*/
\STATE Compute the Jacobian norm $||J(\mathbf{x})||_F^2$; 
\STATE Train the network using Eq. \eqref{eq:loss-ensemble} and update $\theta$.
%\IF {conditional}
%\STATE Perform task A.
%\ELSE
%\STATE Perform task B.
%\ENDIF
\ENDWHILE
\STATE \textbf{return} $\theta$.
\end{algorithmic}
\end{algorithm}

\textbf{Improved robustness using the Jacobian norm:}
In the presence of perturbed examples, the expected response of a DNN should stay similar to the correct prediction, which can be mathematically described as $\Omega = F_{\theta}(\mathbf{x})- F_{\theta}(\hat{\mathbf{x}})$. Suppose $\mathbf{x}_i$ is the $i$-th component of noise-free signal $\mathbf{x}$, and $\hat{\mathbf{x}}_i =\mathbf{x}_i + \epsilon \mathbf{z}_i$ is the noise-crafted tensor variation of $\mathbf{x}$. Note that the term $\Omega$ measures the difference of the predictions in the case of clean data and its perturbed counterpart. $\mathbf{z}$ is the noise term, which is sampled from a Gaussian distribution with zero mean and variance $\sigma^2$ for each inference. Note that the noise term has an identical variance to $\mathbf{x}$ so that the additive noise only relies on the distribution of $\mathbf{x}$ to dynamically perturb the input. According to the above, the linearized robustness can be approximated and upper-bounded by the Jacobian of the classifier's prediction w.r.t the input $\mathbf{x}$. Thus, with the gradient back-propagation, we can determine the gradient calculation w.r.t the difference of the two predictions $\Omega$ via: 
\begin{equation}
\begin{aligned}
&\nabla_{\mathbf{x}} \Omega =\nabla_{\mathbf{x}} ( F_{\theta} (\mathbf{x}) - F_{\theta}(\hat{\mathbf{x}}))= \nabla_{\mathbf{x}}  F_{\theta} (\mathbf{x}) - \nabla_{\mathbf{x}} F_{\theta}(\hat{\mathbf{x}})\\
&\Rightarrow || \nabla_{\mathbf{x}} F_{\theta}(\mathbf{x})- \nabla_{\mathbf{x}} F_{\theta}(\hat{\mathbf{x}}) ||_F^2 \leq || J (\mathbf{x}) ||_F^2.
\end{aligned}
\end{equation}

%To optimize the Jacobian matrix, we incorporate the Jacobian norm as a regularizer into the robust optimization with adversarial training. This is expected to explicitly quantify the most discriminative portion of the image data with a perturbation. Given the input $\mathbf{x}\in \mathbf{X}$ and its target label $\mathbf{y} \in \mathbf{Y}$, we adversarially train the network to have the optimal solution of the network parameters $\theta$ that satisfy the min-max game:
%\begin{equation}\label{eq:min-max}
%\arg\min_{\theta} \left[ \arg \max_{\hat{\mathbf{x}}\in \prod_ {\mathbf{x}+S}} H(F_{\theta}(\hat{\mathbf{x}}), \mathbf{y}) \right],
%\end{equation}
%where the inner maximization intends to acquire the perturbed example $\hat{\mathbf{x}}$ from the perturbed set $\prod_ {\mathbf{x}+S}$. We adopt the $L_{\infty}$ based PGD attack \cite{PGD} as the default inner maximization solver to generate $\hat{\mathbf{x}}$. As stated above, the Jacobian norm regularization for the optimization of DNN under attacks can improve the interpretability of network predictions. However, gradient-generated samples can easily fool other models even though these samples are designed for a specific attack. To combat such transferable examples, we propose to select the most salient input gradients that can imitate the approximation of the DNN in the presence of perturbed data. The proposed selective input gradient regularization will allow the model to efficiently keep high saliency on task-relevant features, and thus maintain the robustness.

The above Jacobian norm can approximate the linear robustness towards the input, which simulates how the network will respond to those small variations of the input. To compute the Jacobian norm, one needs to take the model's gradient with respect to its inputs, which provides a local linear approximation of the model's behavior. However, directly using the raw input gradients is demonstrated to be ineffective since these gradients are quite noisy and hard to interpret \cite{Baehrens-JMLR-2010}. To combat this challenge, in the following we propose to train the classification model under an input gradient regularization with fewer extreme values and a minimized Jacobian norm. 

\subsection{Selective input gradient regularization}

\paragraph{Input gradient regularization} The idea of input gradient regularization was first introduced by Drucker et al \cite{Double-BPP} to train neural networks by minimizing not just the ``energy" of the network but also the rate of the change of the energy with respect to the input features. The energy can be formulated using the cross-entropy as follows:
\begin{equation}\label{eq:input-gradient}
\begin{aligned}
&\theta^*= \arg\min_{\theta} \sum_{n=1}^N \sum_{k=1}^K -\mathbf{Y}_{n,k} \log F_{\theta}(\mathbf{X}_{n,k}) \\
& + \lambda_m \sum_{n=1}^N \sum_{d=1}^D \left(\frac{\partial}{\partial \mathbf{x}_d}\sum_{k=1}^K-\mathbf{Y}_{n,k} \log F_{\theta}(\mathbf{X}_{n,k})\right)^2,
\end{aligned}
\end{equation}
where $\lambda_m$ is a hyperparameter modulating the penalty strength. Such input gradient regularization ensures that even if the input change slightly, the KL divergence between the predictions and the true labels will not be changed significantly. This double back-propagation can provide a constraint on the sensitivity caused by perturbations. Intuitively, the gradient penalty term encourages the predictions not be sensitive to small perturbations in the input space because it regularizes the input gradients to be smoother with fewer extreme values. We combine the Jacobian norm and the input gradient regularization, which can be formulated as follows:
\begin{equation}\label{eq:input-gradient-JN}
\begin{aligned}
&\theta^*= \arg\min_{\theta} \sum_{n=1}^N \sum_{k=1}^K -\mathbf{Y}_{n,k} \log F_{\theta}(\mathbf{X}_{n,k}) \\
& + \lambda_m \sum_{n=1}^N \sum_{d=1}^D \left(\frac{\partial}{\partial \mathbf{x}_d} \sum_{k=1}^K-\mathbf{Y}_{n,k} \log F_{\theta}(\mathbf{X}_{n,k})\right)^2 \\
& + \lambda_j || J (\mathbf{x}) ||_F^2,
\end{aligned}
\end{equation}
where $\lambda_m$ and $\lambda_j$ denote the weights for the selective gradient regularization and the Jacobian normalization, respectively.
However, in Eq. \eqref{eq:input-gradient-JN}, the combination of Jacobian norm and input gradient regularization only provides constraints for very near training examples. Thus, it does not solve the adversarial perturbation problem. It is also expensive to make derivatives smaller to limit the sensitivity to infinitesimal perturbations. With this regard, in the following, we propose a \textit{perturbation based saliency map} to select the most discriminative features, which are not only robust to perturbations but more interpretable to the network behaviour.

\paragraph{Perturbation based saliency for selective input gradient regularization}

Saliency map in deep learning is a technique used to interpret input features that are determined to be important for the neural network output \cite{Deep-Inside,DIGR,SmoothGrad}. As domain experts are more concerned with the interpretability of a DNN, some methods have been proposed to generate saliency maps to explain the decision making in the DNN. One may directly use gradients to estimate the influence of input features on the output. However, the quality of the gradient-based saliency maps is generally poor as gradient-based saliency map methods tend to overly smooth the gradients \cite{Input-gradients-robustness,Saliency-RL}. 

In the spirit of saliency map in highlighting the importance of input features, we propose to use the perturbation based saliency map, denoted as $\mathcal{M}_d =f(\nabla_{\mathbf{x}} F_{\theta}(\hat{\mathbf{x}}))$, which is derived from the gradient of a perturbed input. The $f(\cdot)$ is a mapping function to be detailed later. Such a perturbation-based saliency map can be computed by perturbing the input and observing the change in output, and thus shows high interpretability in a DNN's behaviour. Specifically, we compute this saliency map by resembling the input images and highlighting the most salient parts. Following \cite{JARN}, we use a Generative Adversarial Network (GAN) to generate a saliency map to be visually similar to the real image. Given $J(\hat{\mathbf{x}})=\nabla_{\mathbf{x}} F_{\theta}(\hat{\mathbf{x}})$, we use an aligning network ($f$) to map the Jacobian into the domain of the input image: $J'=f(J(\hat{\mathbf{x}}))$. In our implementation, $f$ (parameterized by $\Phi$) is implemented by a single $1\times1$ convolutional layer with a \texttt{tanh} activation function. Hence, the generator $G(\mathbf{x},\mathbf{y})$ can be framed by using both the classifier and the aligning network: $G_{\theta,\Phi}(\mathbf{x},\mathbf{y})=f(\nabla_{\mathbf{x}} F_{\theta}(\hat{\mathbf{x}}))$. As a result, the generator can learn to model the distribution of $p_{J'}$ to resemble that of $p_{\mathbf{x}}$.

Once $\mathcal{M}_d$ is obtained, we incorporate the salient map into input gradient computation such that the most robust gradients can be selected during back-propagation. Mathematically, the objective with Jacobian norm and selective input gradient regularization is defined as
\begin{equation}\label{eq:loss-ensemble}
\begin{aligned}
&\theta^*= \arg\min_{\theta} \sum_{n=1}^N \sum_{k=1}^K -\mathbf{Y}_{n,k} \log F_{\theta}(\mathbf{X}_{n,k}) \\
& + \lambda_m \sum_{n=1}^N \sum_{d=1}^D \left(\frac{\partial}{\partial \mathbf{x}_d} \mathbbm{1} (\beta-\mathcal{M}_d) \sum_{k=1}^K-\mathbf{Y}_{n,k} \log F_{\theta}(\mathbf{X}_{n,k})\right)^2 \\
& + \lambda_j || J (\mathbf{x}) ||_F^2.
\end{aligned}
\end{equation}

We suggest that the two terms have equal importance to the overall optimization, and thus set $\lambda_m=\lambda_j=0.5$ as default configuration except for otherwise specified. The term $|| J (\mathbf{x}) ||_F^2$, i.e., the Jacobian Frobenius norm acts as a regularizer clipping the values of inputs such that the gradients of classification logits with respect to the inputs can ensure the linearized robustness in the presence of perturbed data, and thus we can minimize the number of mis-predictions of the learned classifier. However, this is ineffective by directly evaluating each raw input features, which could be very noisy. Then, the input gradient regularization, as being modulated by $\lambda_m$, can smooth the gradients to be less noisy with fewer extreme values. Finally, in Eq.\eqref{eq:loss-ensemble} we embed the perturbation-based saliency map ($\mathcal{M}_d$) to improve the interpretability of a neural network's prediction. The indicator function $\mathbbm{1}$ is to determine whether the saliency for an input feature is below a threshold $\beta$, and thus it returns 1 if $\beta-\mathcal{M}_d \geq 0$ and 0 otherwise. The whole training procedure of the proposed method is illustrated in Algorithm \ref{alg:algorithm-J-SIGR}.

\section{Experiments}\label{sec:experiment}
To evaluate the robustness of the proposed J-SIGR, we conducted experiments on two image datasets: MNIST \cite{MNIST} and CIFAR-10 \cite{CIFAR-10}. Below, we first describe the experimental settings and then report the experimental results under a range of attacks. Finally, we performed ablation studies to provide a more insightful analysis of the proposed method.

\subsection{Experimental settings}

\paragraph{Datasets}
1) The MNIST dataset \cite{MNIST} consists of handwritten digit $28\times 28$ gray-scale images that are divided into 60K training and 10K test images. We trained a CNN, composed of 3 convolutional layers and one final soft-max layer, to suit the small-sized MNIST. All convolutional blocks have a stride of 5 while each layer has an increasing number of output channels (i.e., $c$= 64-128-256). 2) The CIFAR-10 \cite{CIFAR-10} dataset contains a collection of $32\times 32\times 3$ colored images that are categorized into 10 classes with 50K training and 10K test images. We use the ResNet-20 architecture \cite{ResNet} with 20 convolutional layers to train the images from CIFAR-10. Throughout the network, the kernel size is set to $9\times9$ in all convolutional layers and the number of channels is increased from 9, 18 to 36 for the three building blocks, respectively.

\paragraph{Implementations and evaluation metrics} ResNet-20 architecture \cite{ResNet} was used as the backbone for most of the comparative experiments and ablation studies. We adopted the data augmentation \cite{ResNet,PNI} but without the input normalization. Alternatively, we placed a non-trainable data normalization layer preceding the DNN to perform the identical function so that the attack tactics can directly add perturbations into the natural images. Since our method involves randomness, we reported the accuracy in the format of mean$\pm$ std with 10 trails to compute the statistical values.

To measure the robustness of both white and black-box attacks, we tested all models against adversarial examples generated for each model and reported the accuracy. On the JSMA setting \cite{JSMA}, where the generated adversarial examples would resemble the targets rather than their original labels, we followed \cite{Input-gradients-robustness} to adopt a human subject experiment to evaluate the legitimacy of adversarial example misclassifications.

\paragraph{Attacks}
To thoroughly evaluate the performance of our proposed method, we considered multiple powerful attacks. The white-box attacks include FGSM \cite{FGSM} and PGD \cite{PGD}. FGSM \cite{FGSM} is an efficient single-step adversarial attack scheme. Given a vectorized input $\mathbf{x}$ and its target label $\mathbf{y}$, FGSM \cite{FGSM} alters each element of $\mathbf{x}$ along the direction of its gradient with respect to the inference $\partial F_{\theta}(\mathbf{x})/\partial \mathbf{x}$. The PGD attack \cite{PGD}, known as one of the strongest $L_{\infty}$ adversarial example generation algorithms, is a multi-step variant of FGSM \cite{FGSM}. The iterative update of crafted data $\hat{\mathbf{x}}$ in the $t+1$-th step can be expressed as $\hat{\mathbf{x}}^{t+1}=\prod_{\mathbf{x}+S}(\hat{\mathbf{x}}^t + \epsilon \cdot \mbox{sgn} (\nabla_{\mathbf{x}} (F_{\theta}(\hat{\mathbf{x}}^t), \mathbf{y})))$, where $\prod$ is the projection space bounded by $\mathbf{x}\pm S$, and $\epsilon$ is the step size. For the PGD attack \cite{PGD} on two datasets, $S$ is set to 0.3/1 and 8/255, and the number of iterations $N_{step}$ is set to 40 and 7, respectively. FGSM \cite{FGSM} adopts the same $\epsilon$ setup as PGD \cite{PGD}. The attack configurations of PGD \cite{PGD} and FGSM \cite{FGSM} were chosen identical to the setup in many adversarial defense methods \cite{PNI,PGD}. To generate adversarial examples for JSMA \cite{JSMA}, we used the Cleverhans adversarial example library \cite{Cleverhans}. We also evaluated the proposed method against several state-of-the-art black-box transferred attacks: substitute \cite{Substitute}, ZOO \cite{ZOO} and transferable attacks \cite{Transfer-attack}. 

\paragraph{Defenses}

To evaluate the improved robustness of our method, we compared it with state-of-the-art defense models: adversarial training \cite{Adversarial-features}, distillation \cite{Distillation-Defense} and a gradient regularization based model \cite{Input-gradients-robustness}. More specifically, for adversarial training, we trained FGSM \cite{FGSM} with perturbations at $\epsilon=0.3$. For distillation based defense, we used a soft-max of temperature $T=50$. For the gradient regularization based model as shown in Eq. \eqref{eq:input-gradient}, we used the double back-propagation to train the classification model.

\subsection{Robustness under FGSM and PGD attacks}

\begin{figure*}[t]
  \centering
  \includegraphics[width=1\linewidth]{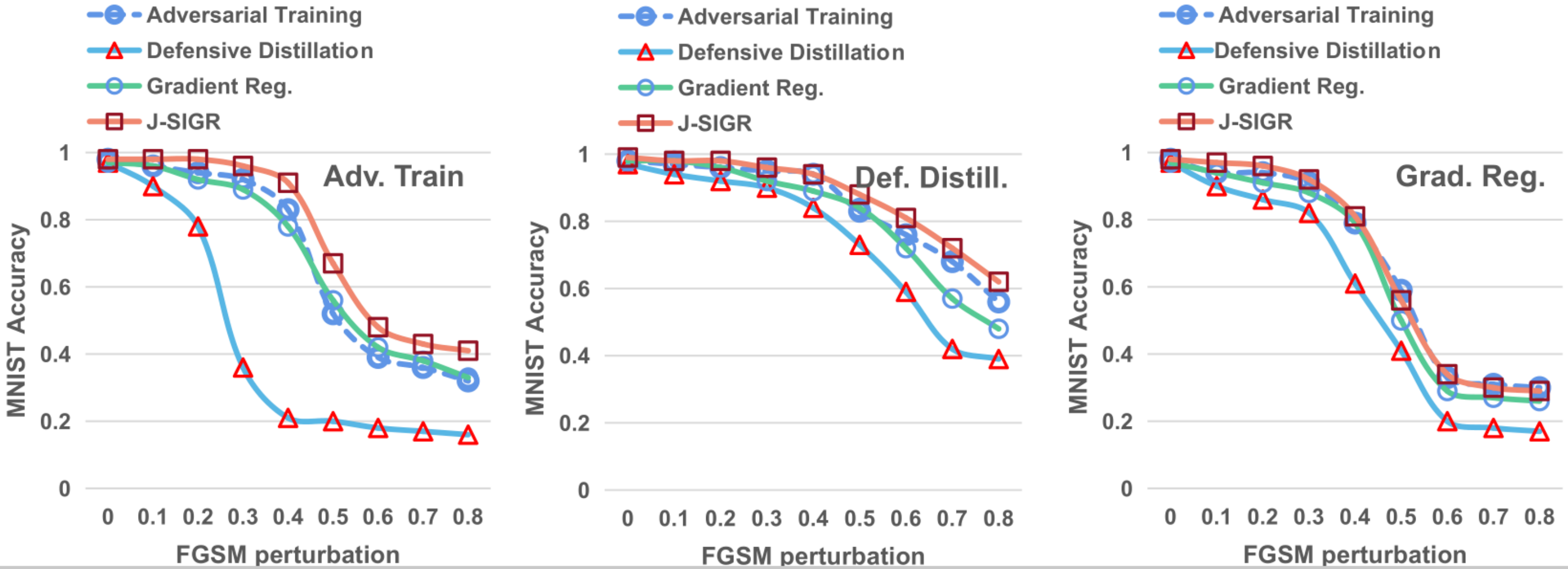}\\
  \vspace{0.5cm}
  \includegraphics[width=1\linewidth]{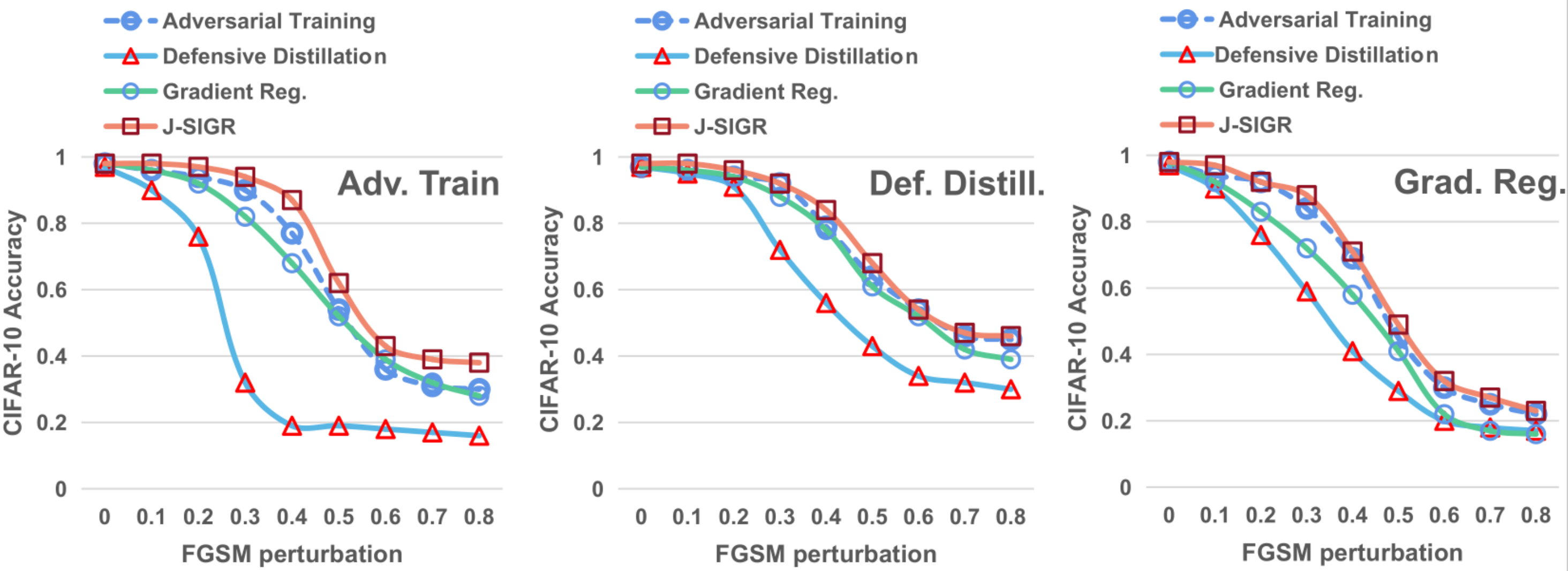}
  \caption{Model accuracy for FGSM examples generated to fool adversarial training, defensive distillation, and gradient regularization based models (from left to right).}\label{fig:FGSM-attack-robustness}
\end{figure*}

\begin{figure}[t]
  \centering
  \includegraphics[width=1\linewidth]{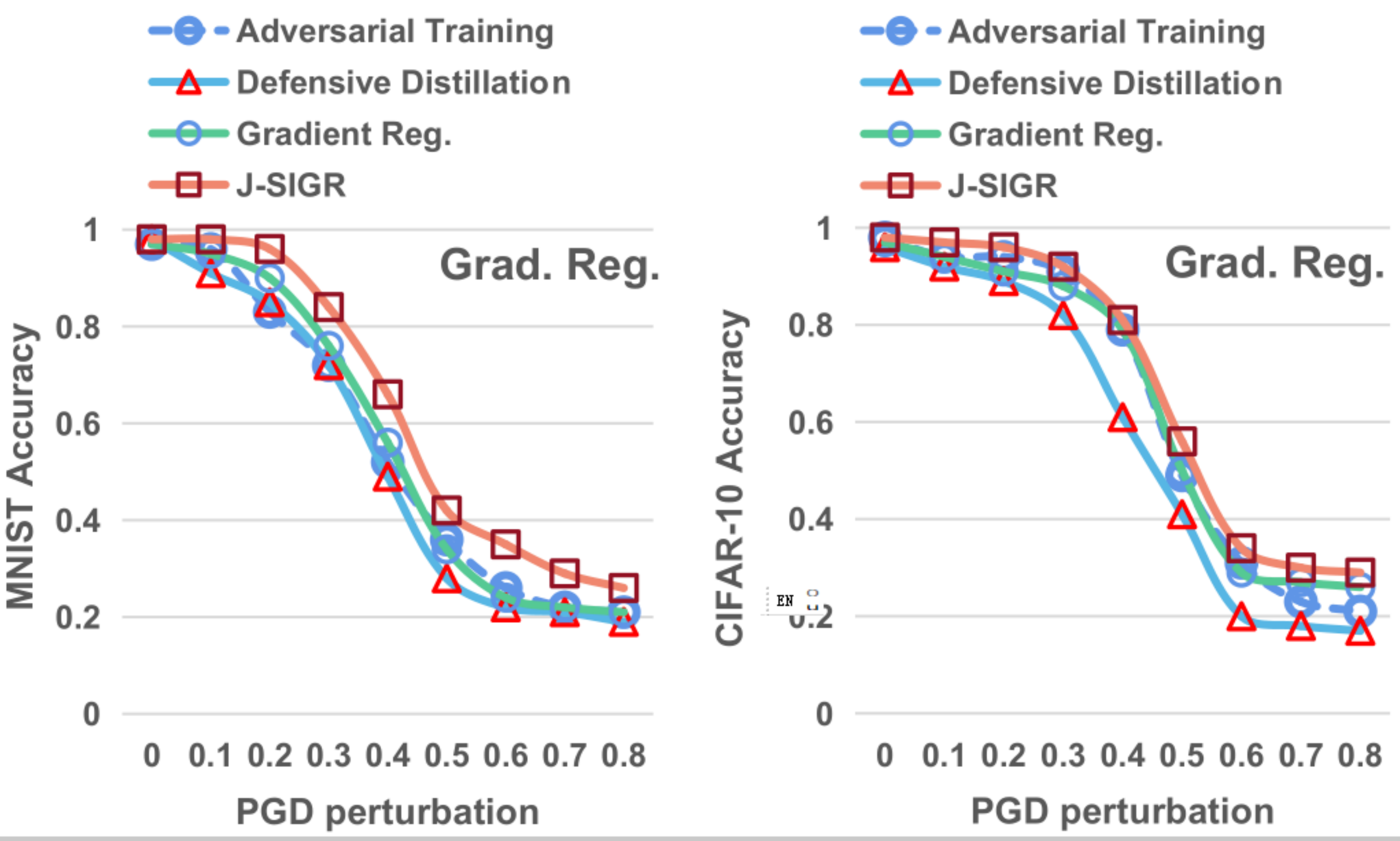}\\
  \caption{Defensive model accuracy against PGD attack when applying gradient regularization as the fool target.}\label{fig:PGD-attack-robustness}
\end{figure}

In Fig. \ref{fig:FGSM-attack-robustness}, we show the robustness results of our method as well as the performance of other defensive models under the FGSM attack \cite{FGSM} on two datasets. It can be observed that the gradient-regularized model \cite{Input-gradients-robustness} exhibits strong robustness to transferred FGSM \cite{FGSM} attack (examples produced by attacking other models). For example, on the MNIST dataset, the adversarial examples produced by attacking the defensive distillation can successfully fool the model based on adversarial training. In contrast, the gradient regularization based methods (including the proposed J-SIGR) can still maintain the accuracy. We evaluate the robustness of the gradient regularization models under a different attack, i.e., PGD \cite{PGD}, and report the results in Fig. \ref{fig:PGD-attack-robustness} on two datasets. Under this attack, the adversarial examples are generated to fool a gradient regularized model, while the results of the two models show that gradient regularization is still effectively robust against a white-box attack \cite{PGD}. Interestingly, gradient-regularized models seem to be vulnerable to white-box attacks, but can still fool all other models. In this respect, in the presence of adversarially transferred examples, we hypothesize that gradient regularization is particular not only for defense but also attack.

\begin{table}[t]
\caption{ Convergence of gradient regularization with layer-wise Jacobian norm (LW-JN) on the CIFAR-10 dataset with ResNet-20 as backbone. The Jacobian norm is given for lower convolutional layers (Conv1 to Conv$3_5$). Test accuracy for perturbed data is computed for the PGD and FGSM attacks.}
\begin{center}
\begin{tabular}{|c|c|c|c|}
\hline
Layer & Vanilla Train & JN+AT & Grad.Reg.+LW-JN\\
\hline
Conv1 &0.004 &0.157& 0.155\\
\hline
Conv$2_0$ & 0.003 & 0.089 & 0.091\\
Conv$2_1$ & 0.001 & 0.067 & 0.061\\
Conv$2_2$ & 0.002 & 0.055 & 0.057 \\
Conv$2_3$ & 0.002 & 0.099 & 0.098 \\
Conv$2_4$ & 0.004 & 0.782 & 0.580 \\
Conv$2_5$ & 0.004 & 0.422 & 0.333 \\
\hline
Conv$3_0$ & 0.002 & 0.087 & 0.079\\
Conv$3_1$ & 0.000 & 0.064 & 0.064\\
Conv$3_2$ & 0.003 & 0.072 & 0.069\\
Conv$3_3$ & 0.001 & 0.062 & 0.060\\
Conv$3_4$ & 0.001 & 0.046 & 0.046\\
Conv$3_5$ & 0.000 & 0.022 & 0.021\\
\hline
FC & 0.001 & 0.013 & 0.012 \\
\hline
PGD & 0.00 & 0.54 $\pm$ 0.11 & 0.57$\pm$0.10\\
FGSM & 0.14 & 0.62 $\pm$ 0.10 & 0.66$\pm$0.09\\
\hline
\end{tabular}
\end{center}
\label{tab:white-attack}
\end{table}

\begin{figure}[t]
  \centering
  \includegraphics[width=0.8\linewidth]{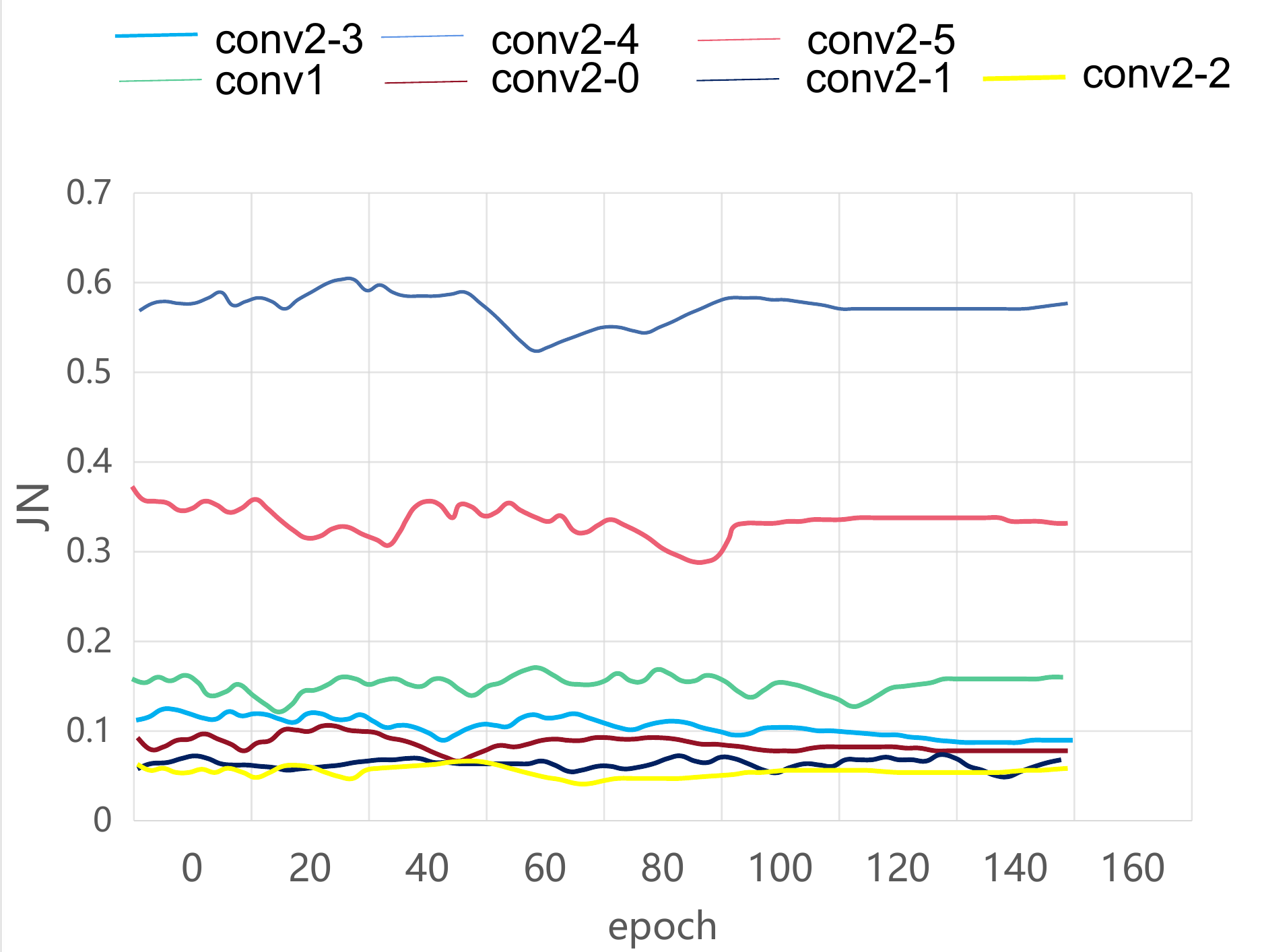}
  \caption{Evolution curves of the Jacobian norm computed at each convolutional layer. Best viewed in color.}\label{fig:JN-epoch}
\end{figure}

Since our model consists of two robustness mechanisms, we investigated the impact of Jacobian norm (JN) by disabling double back-propagation and examining the output response of each layer with respect to two different attacks. More specifically, a Jacobian-norm based variant of our method was implemented by adding layer-wise Jacobian norm into the DNN, together with the input gradient regularization. This variant is called \textit{Grad.Reg.+LW-JN}. As shown in Table \ref{tab:white-attack}, only performing vanilla training using momentum SGD optimizer can lead to the failure of adversarial defense with the values of Jacobian norm converging towards negligible values. After applying the JN as a regularization of the network (i.e., JN+AT), the lower convolutional layers attain a relatively large JN. The variant of our method with layer-wise Jacobian norm (i.e., \textit{Grad.Reg.+LW-JN}) achieves the highest performance with respect to the two attacks. This demonstrates robustness improvement by leveraging the proposed network architecture, which is parameterized to resist perturbations via gradient back-propagation. Since JN can reflect the robustness of the network, we plotted the evolution curves of the JN values for the lower convolutional layers to illustrate the robustness of the network connections (Fig. \ref{fig:JN-epoch}).

\subsection{Robustness under black-box attacks}

In this experiment, we evaluate our scheme against three black-box attacks: transferable adversarial attack \cite{Transfer-attack}, Substitute \cite{Substitute} and ZOO attack \cite{ZOO}. Following the setting of \cite{Transfer-attack}, two neural networks were trained with their individual architectures with one network chosen as the source model and the other chosen as the target model. An adversarial example $\hat{\mathbf{x}}$ generated from the source model was then used to attack the target model without access to the parameters of the target model. This is denoted as $Source \rightarrow Target$. We trained two ResNet-18 models (i.e., $M_1$ and $M_2$) on CIFAR-10 dataset to attack each other with $M_1$ optimized through PGD adversarial training and $M_2$ optimized through our proposed method. Experimental results are given in Table \ref{tab:transferred}. Our proposed method achieves higher accuracy under two transferable attacks and is seen similar perturbed-data accuracy for both transferable scenarios. This indicates that our method provides robustness against transferable black-box attack. This also shows that the presence of J-SIGR has negligible effect on inference under a strong attack PGD. For the other two types of attacks, we evaluated our defense ability on 200 randomly selected test examples for an untargeted attack. The success rate refers here to the percentage of test samples which are wrongly classified under the attack. For example, the ZOO attack success rate for vanilla ResNet-18 with adversarial training is close to 80\%. The results in Table \ref{tab:transferred} suggest that our method is robust as it resists the two attacks by noticeably dropping the success rate from 80\% to 49\% and 77\% to 48.8\% under ZOO attack \cite{ZOO} and Substitute \cite{Substitute}, respectively (Lower success rates means higher robustness).

\begin{table}[t]
\centering
\caption{The proposed method (J-SIGR) against transferred attacks on CIFAR-10 test set. Model $M_1$ is trained by PGD-AT based on ResNet-18 architecture and $M_2$ is trained on ResNet-18 using our method (J-SIGR).}\label{tab:transferred}
\begin{center}
\begin{tabular}{cc|c|c}
    \hline
     \multicolumn{2}{c|}{Transferable attack} & ZOO & Substitute\\
    \cline{3-4}
     $M_1 \rightarrow M_2$ & $M2 \rightarrow M_1$ & success rate (80\%) & success rate (77\%) \\
     \hline
     $78.14 \pm 0.26$ & $76.82\pm 0.19$ & 49.00 & 48.80\\
    \hline
\end{tabular}
\end{center}
\end{table}

\begin{figure}[t]
  \centering
  \includegraphics[width=0.8\linewidth]{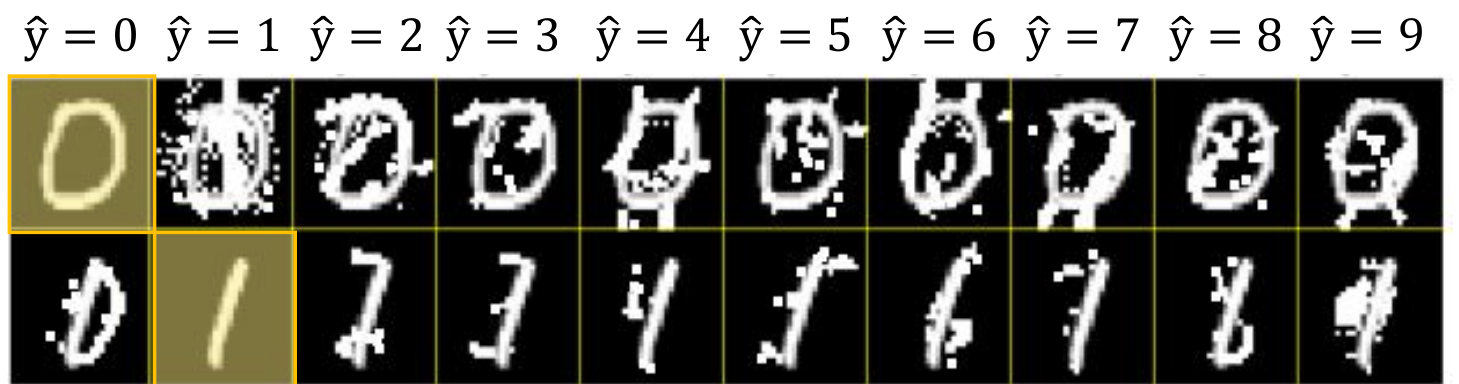}
  \caption{Perturbations by applying JSMA to digits 0 and 1 with maximum distortion parameter $\gamma=0.25$ for a gradient regularization model. The highlighted images in each row are modified until the model predicts the digit corresponding to their column or the maximum distortion is reached.}\label{fig:JSMA-MNIST}
\end{figure}

\subsection{Evaluation on human subject study under Jacobian-saliency map attack (JSMA) }

Unlike other attacks that stop generating adversarial examples when the maximum distortion is met, JSMA \cite{JSMA} constantly generates adversarial examples until the model predicts the target. Thus, evaluating the robustness under JSMA \cite{JSMA} using accuracy numbers is not appropriate. This is also because the perturbations created by JSMA alter the adversarial examples so they resemble the target labels instead of the original labels. As shown in Fig. \ref{fig:JSMA-MNIST}, for a gradient regularized model, we applied JSMA on each image 0 or 1 to generate perturbations until the model predicts the digit corresponding to their column target label or the maximum distortion is reached (We set the maximum distortion $r=0.25$). Then, we followed \cite{Input-gradients-robustness} to test these different robustness scenarios using 11 human subjects who were invited to evaluate whether examples generated by different methods are more or less plausible instances of their targets. Specifically, the subjects were shown 30 images of JSMA-crafted MNIST examples, with each of these 30 images corresponding to one original digit (from 0 to 9) and one model (defensive distillation, gradient regularization and selective input gradient regularization). Images were randomly and uniformly sampled from a larger set of 45 examples corresponding to the first 5 images of the original digit in the test set. Images in the test set were transformed by using JSMA to resemble each of the other 9 digits. Subjects were not provided the original labels and were asked to identify the most two plausible predictions for the image they believed a classifier would produce (they entered N/A if they found no label was a good choice). 

Table \ref{tab:human-subjects} shows the quantitative results from the human subject experiment. Overall, human subjects found that gradient-regularized models can generate the most convincing examples to fool humans. More specifically, humans mostly believe gradient-regularized adversarial examples (both the input gradient regularization \cite{Input-gradients-robustness} and our method) are favourably classified as their target labels instead of the original digits. For example, the values of ``human fooled" column in Table \ref{tab:human-subjects} show that the mispredictions of gradient regularized models are very ``reasonable" in comparison to adversarial training and defensive distillation.

\begin{table}[t]
\centering
\caption{Quantitative results form human subject experiment on MNIST dataset. SIGR stands for selective input gradient regularization. The measure``human fooled" records the percentage of examples which are classified by human subjects as the most plausibly adversarial targets (the higher, the better). ``Mistake reasonable" measures the percentage of examples which are classified as either the target plausible or unrecognizable as any label (the higher, the better). Best results are in bold.}\label{tab:human-subjects}
\begin{center}
\begin{tabular}{|c|c|c|}
    \hline
     & \multicolumn{2}{c|}{MNIST (JSMA)} \\
    \hline
     Model & human fooled & mistake reasonable \\
     \hline
     Def. Distill & 0.0\% & 23.5\% \\
     Grad. Reg. & 16.4\% & 41.8\% \\
     SIGR & \textbf{20.2}\% & \textbf{45.1}\%\\
    \hline
\end{tabular}
\end{center}
\end{table}

\subsection{Comparison with other defensive models for both clean and crafted data}

Most adversarial defense methods focus on corrupted data, while neglecting clean data. However, a good defensive model should perform well in the presence of both clean and corrupted data. We thus compared our method with state-of-the-art methods by evaluating performance for both clean and crafted data. 

Experimental results are reported in Table \ref{tab:compare-SOTA}. In previous experiments, we compared against different defense models that are to date still unbroken including the PGD based adversarial training \cite{PGD} and several randomness-based works \cite{DP,PNI,RSE,Adv-BNN}. Additionally, we compared against JARN \cite{JARN}, which utilizes the Jacobians to generate images resembling to the original images. The compared performance results are shown in Table \ref{tab:compare-SOTA}. Note that previous defense efforts \cite{Thermometer} often achieve improved accuracy on contaminated data at the expense of lowering clean data accuracy. In contrast, we introduced a notion of linearized robustness which performs well in both clean and perturbed data. As shown in Table \ref{tab:compare-SOTA}, in comparison to the PGD-based adversarial training, our proposed method outperforms all methods for both clean and perturbed data accuracy under the white-box attack. For example, differential privacy (DP) \cite{DP}, which introduces noises at various locations of the network so as to guarantee a certified defense, does not perform well against $L_{\infty}$-norm based attacks, e.g., PGD \cite{PGD} and FGSM \cite{FGSM}. Moreover, to pursue a higher level of certified defense, DP dramatically reduces the clean data accuracy down to 25.1\%. A set of noise injection methods, i.e., RSE \cite{RSE}, Adv-BNN \cite{Adv-BNN} and PNI \cite{PNI}, combine the adversarial training and noise injection into the inputs/weights of the network. However, these methods, except for the PNI \cite{PNI} manually set the noise configurations, making it very ad-hoc, and thus not generalizable to different datasets. PNI \cite{PNI} exploits the min-max optimization with trade-off on clean-and perturbed data by injecting trainable Gaussian noise on various locations of the network to generate adversarial examples. Whilst PNI \cite{PNI} improves the accuracy of both clean and perturbed data, noise injection is not related to the robustness response of the network. In contrast, our proposed method regularizes the Jacobian norm and the input gradients, such that the network parameters can be dynamically trained to perform better adversarial defense. In addition, the Jacobian norm regularization explicitly suggests the robustness of the classification model in response to imperceptible data perturbation.

\begin{table}[t]
\centering
\caption{Comparison with state-of-the-art defense methods using clean and perturbed data accuracies on CIFAR-10 under PGD attack.}\label{tab:compare-SOTA}
\begin{tabular}{rcc}
\hline
Defense method & Clean & PGD\\
\hline
PGD-AT \cite{PGD}  & 87.0 $\pm$ 0.1 & 46.1 $\pm$ 0.1\\
DP \cite{DP}  & 87.0 & 25.1 \\
RSE \cite{RSE} & 87.5 & 40.0 \\
Adv-BNN \cite{Adv-BNN} & 79.7 & 45.4 \\
PNI \cite{PNI} & 87.1$\pm$ 0.1& 49.1$\pm$0.3\\
JARN \cite{JARN} & 84.8 & 51.8 \\
J-SIGR (Ours) & 90.1 $\pm$ 0.2& 57.6 $\pm$ 0.4 \\
\hline
\end{tabular}
\end{table}

\begin{figure}[t]
  \centering
  \includegraphics[width=0.8\linewidth]{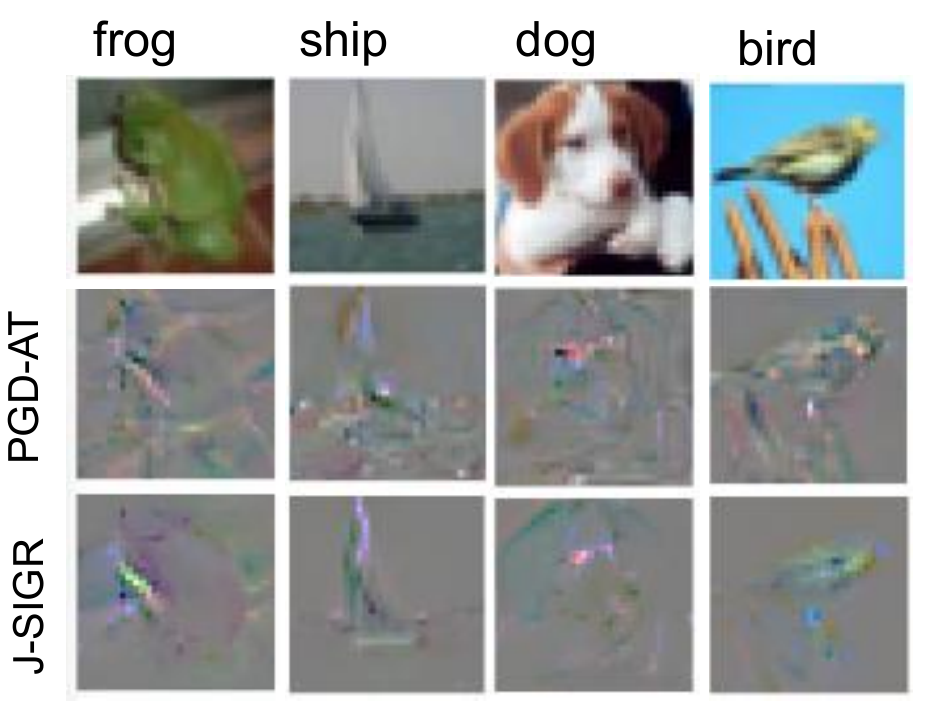}
  \caption{Visualization of Jacobian matrix of models (PGD-AT and our method) on CIFAR-10 dataset.}\label{fig:Jacobian-matrix}
\end{figure}

\begin{figure}[t]
  \centering
  \includegraphics[width=0.8\linewidth]{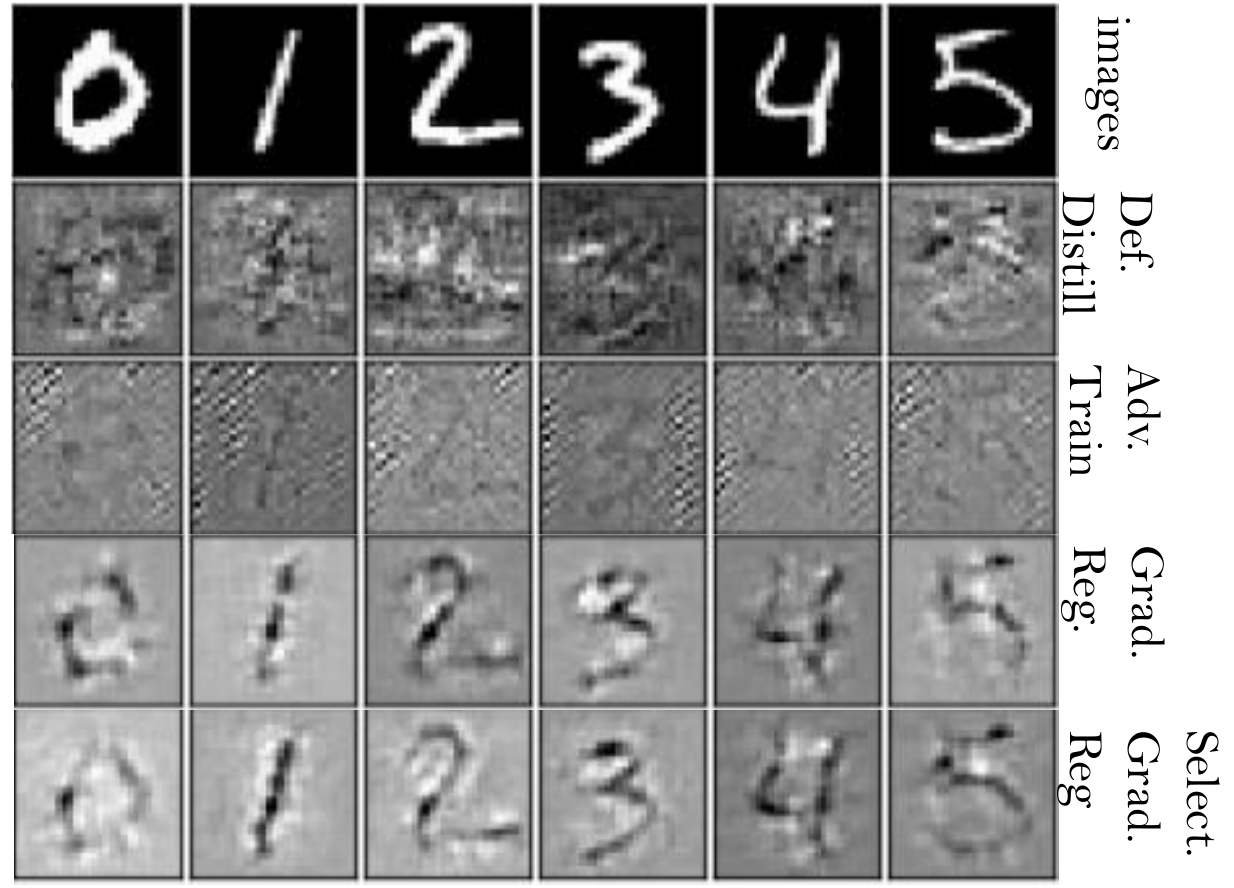}
  \caption{Visualization of input gradients on MNIST dataset.}\label{fig:input-gradients}
\end{figure}

\begin{figure}[t]
  \centering
  \includegraphics[width=0.8\linewidth]{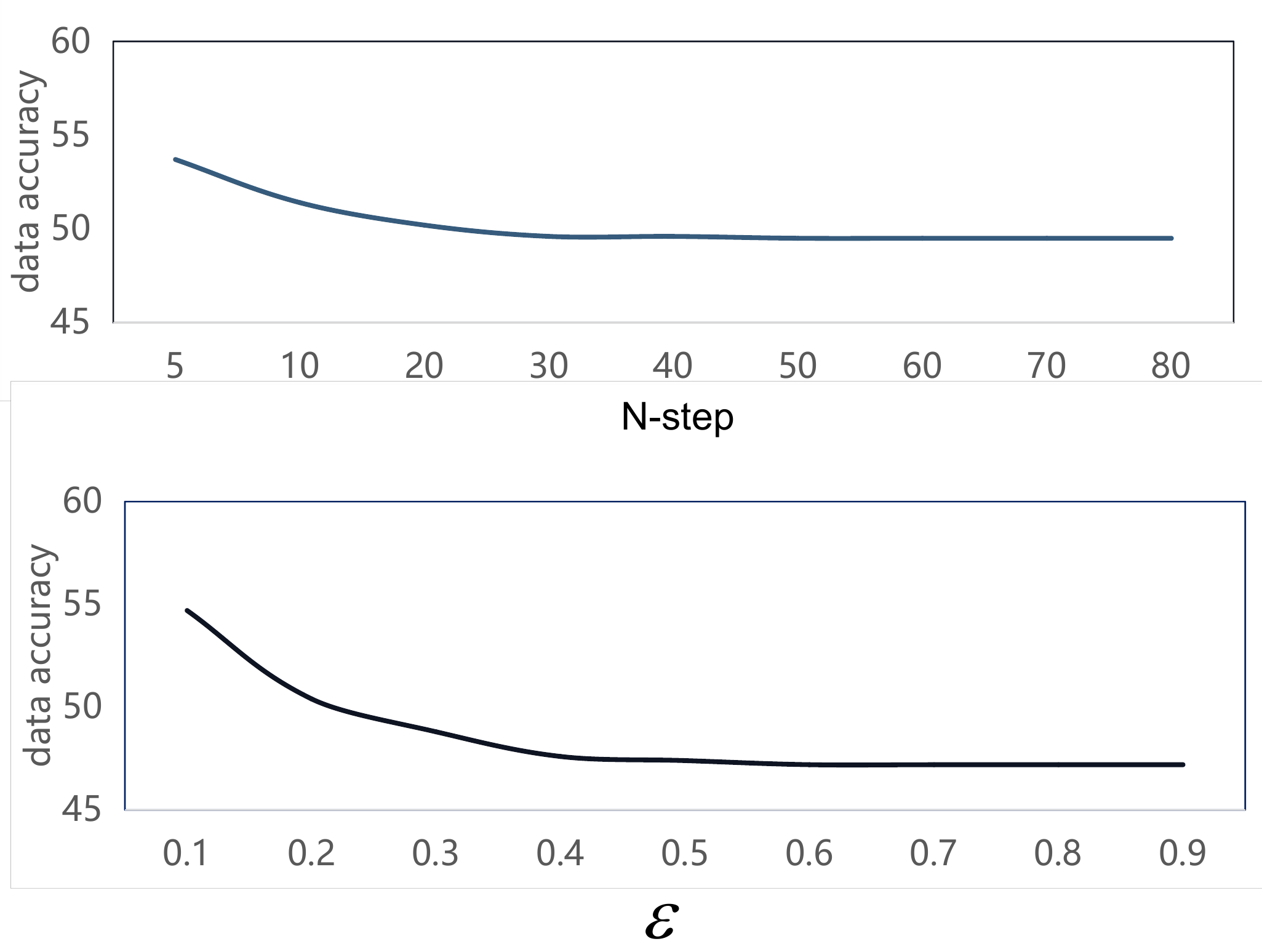}
  \caption{Perturbed data accuracy of CIFAR-10 test set under the PGD attack versus number of attack steps N-step (top) and attack bound $\varepsilon$ (bottom).}\label{fig:accuracy-step-attack-bound}
\end{figure}

\subsection{Connections to network interpretability}

\paragraph{Understanding the Jacobian matrices} An adversarially trained model can gain robustness and also produce salient Jacobian matrices as byproduct. It has been shown that the saliency in Jacobians is a result of robustness \cite{Robust-jacobian}. Thus, it is interesting to use the Jacobian saliency to evaluate how robust a model is. In this study, we visualize the Jacobian matrices of the proposed model and an adversarial-trained PGD \cite{PGD} to show how salient the Jacobian map is. As shown in Fig. \ref{fig:Jacobian-matrix}, the proposed method can better visually resemble the corresponding images than PGD-AT. This demonstrates the improved robustness of the proposed method. 

\paragraph{Understanding the input gradients}
Fig. \ref{fig:input-gradients} visualizes the input gradients across different defensive models on the MNIST dataset. This qualitative visualization shows the different interpretability of the input gradients derived from models based on defensive distillation, adversarial training, gradient regularization and the proposed selective gradient regularization. The adversarially trained model can provide more interpretable gradients than defensive distillation, but not as highly interpretable as gradient regularized models. The proposed method presents the most interpretable gradients, and thus can provide an explanation for adversarial attacks.

\begin{table*}[t]
\caption{ The proposed method (J-SIGR) with various architectures on CIFAR-10 test set. Model $M_1$ is trained by a PGD based adversarial training (PGD-AI) with ResNet-18 as backbone, and $M_2$ is trained on ResNet-18 by our method with AT.}
\begin{center}
\begin{tabular}{cccccccccc}
    \hline
    & \multicolumn{3}{c}{No defense} & \multicolumn{3}{c}{Vanilla Adv. Train} & \multicolumn{3}{c}{J-SIGR}\\
    \cline{2-10}
     Model & Clean & PGD & FGSM & Clean & PGD & FGSM & Clean & PGD & FGSM \\
     \hline
     Net20 & 92.1 & 0.0$\pm$0.0 & 14.1 & 83.8 & 39.1$\pm$0.1 & 46.4 & 90.1$\pm$0.2 & 53.7$\pm$0.3 & 57.6$\pm$0.1 \\
     \hline
     Net32 & 92.8 & 0.0$\pm$0.0 & 17.8 & 85.6 & 42.1$\pm$0.0 & 50.3 & 91.1$\pm$0.2 & 52.8$\pm$0.1 & 54.2$\pm$0.1 \\
     Net44 & 93.1 & 0.0$\pm$0.0 & 23.9 & 85.9 & 40.8$\pm$0.1 & 48.2 & 90.0$\pm$0.1 & \textbf{55.4}$\pm$0.1 & \textbf{58.6}$\pm$0.2 \\
     Net56 & 93.3 & 0.0$\pm$0.0 & 24.2 & 86.5 & 40.1$\pm$0.1 & 48.8 & 92.1$\pm$0.2 & 54.9$\pm$0.2 & 55.8$\pm$0.1 \\
    \hline
     Net20 (1.5$\times$) & 93.5 & 0.0$\pm$0.0 & 15.9 & 85.8 & 42.1$\pm$0.0 & 49.6 & 91.4$\pm$0.1 & 55.2$\pm$0.3 & 55.8$\pm$0.1 \\
     Net20 (2$\times$) & 94.0 & 0.0$\pm$0.0 & 13.0 & 86.3 & 43.1$\pm$0.1 & 52.6 & 91.5$\pm$0.1 & 55.1$\pm$0.2 & 55.0 $\pm$0.1 \\
    \hline
\end{tabular}
\end{center}
\label{tab:effect-depth-width}
\end{table*}

\subsection{Ablation studies}

As discussed in Section \ref{sec:method}, the Jacobian norm of the network output with respect to the input relates to the linearized robustness of the DNN. Our proposed technique introduces tight bounds to the response of the output layer to adversarial perturbation added to the input. Herein, we raise two concerns in regards to our proposed regularization term: 1) whether the robustness improvement introduced by our proposed method is not relying on the stochastic gradient; 2) how the scale of the network architecture (i.e., width and depth) affects the robustness of the DNN. Our first evaluation aims to show that our method is free of gradient obfuscation \cite{Gradient-Obfuscation} by increasing the PGD \cite{PGD} attack steps and the attack bound $\varepsilon$.

\paragraph{Influence of the network capacity}  In order to investigate the links between the network capacity (i.e., width, depth and number of trainable parameters) and the robustness improvement via J-SIGR, we analyzed various network architectures in terms of depth and width. For varied depth, we considered ResNet20/32/44/56 and conducted experiments under vanilla training \cite{PGD} and our technique. For varied width, we employed the original ResNet-20 as the baseline and expanded the input/output channel of each layer by 1.5$\times$ and 2$\times$ scale, respectively. We report both clean and perturbed data accuracies using the network trained with Jacobian term. The results in Table \ref{tab:effect-depth-width} suggest that increasing the model's capacity positively improves the network robustness against white-box attacks, and our proposed method outperforms vanilla training in both clean and perturbed data accuracy for powerful PGD and FGSM attacks. The other observation is that the noticeable robustness improvement provided by our method is indeed provided by the effective training with the proposed J-SIGR. Our method updates the network parameters without introducing randomness in the test phase.

\begin{figure}[t]
  \centering
  \includegraphics[width=0.5\linewidth]{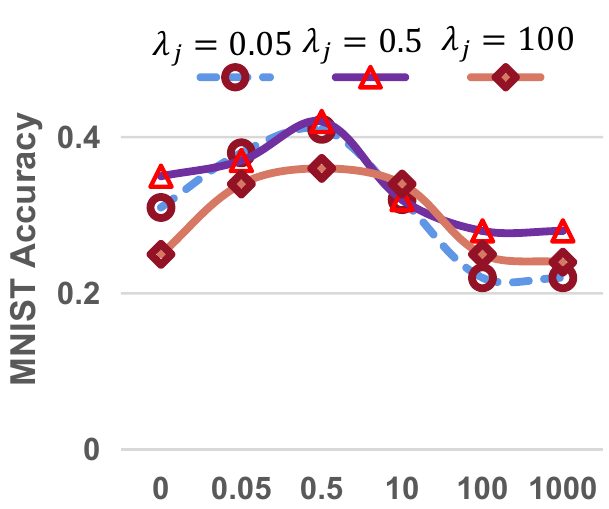}
  \caption{The hyperparameter study on MNIST dataset.}\label{fig:lambda}
\end{figure}

\paragraph{Effect of hyperparameters} In this experiment, we study the effect of two hyperparameters, i.e., $\lambda_m$ and $\lambda_j$ on the MNIST accuracy under FGSM. To de-correlate the impact of two parameters, we fix $\lambda_m=0.5$ and examine the model accuracy with varied value of $\lambda_j$. The results are reported in Fig. \ref{fig:lambda}. As shown in Fig. \ref{fig:lambda}, the highest accuracy of DNN under an attack is achieved when we set $\lambda_j=0.5$. Thus, we empirically $\lambda_m=\lambda_j=0.5$ in all experiments.

\paragraph{Non-dependence on stochastic gradients} To prove that the robustness improvement provided by our method is not due to stochastic gradients, we examine the perturbed data accuracy by evaluating the PGD attack steps N-step and the attack bound $\varepsilon$. As shown in Fig. \ref{fig:accuracy-step-attack-bound}, increasing the attack steps or attack bound can boost the attack strength, which inevitably leads to accuracy degradation. However, the accuracy does not degrade further when N-step=40 or $\varepsilon\geq 0.5$. If the stochastic gradient was improving robustness, then increasing the attack strength would have broken the defense of our method. This is not observed in reported experiments.

\section{Conclusion}\label{sec:con}

In this paper, we proposed an approach based on Jacobian normalization and selective input gradient regularization (dubbed J-SIGR) to improve both robustness and high interpretability of deep neural networks (DNNs). Our method employs the Jacobian matrices to generate gradient-based salient maps, which select informative input gradients to achieve efficient interpretability of adversarial perturbations.
We performed ample experiments to show that our method can improve the robustness of DNNs under multiple adversarial attacks. We believe our approach could help build trustworthy real-world systems and benefit the deployment of deep learning in practice.

%\section*{Acknowledgement}
%This work was funded by Australian Research Council (Grants DP210101682 and DP210102674). This work was partially funded by Anhui Natural Science Foundation Anhui Energy Internet Joint Fund (No. 2008085UD07), Anhui Provincial Key Research and Development Project (No. 202104a07020029), and Natural Science Foundation of China (No. 61876002).

\ifCLASSOPTIONcaptionsoff
  \newpage
\fi



\bibliographystyle{IEEEtran}\small
\bibliography{allbib}

\begin{thebibliography}{10}
\providecommand{\url}[1]{#1}
\csname url@samestyle\endcsname
\providecommand{\newblock}{\relax}
\providecommand{\bibinfo}[2]{#2}
\providecommand{\BIBentrySTDinterwordspacing}{\spaceskip=0pt\relax}
\providecommand{\BIBentryALTinterwordstretchfactor}{4}
\providecommand{\BIBentryALTinterwordspacing}{\spaceskip=\fontdimen2\font plus
\BIBentryALTinterwordstretchfactor\fontdimen3\font minus
  \fontdimen4\font\relax}
\providecommand{\BIBforeignlanguage}[2]{{%
\expandafter\ifx\csname l@#1\endcsname\relax
\typeout{** WARNING: IEEEtran.bst: No hyphenation pattern has been}%
\typeout{** loaded for the language `#1'. Using the pattern for}%
\typeout{** the default language instead.}%
\else
\language=\csname l@#1\endcsname
\fi
#2}}
\providecommand{\BIBdecl}{\relax}
\BIBdecl

\bibitem{Crime-prediction}
H.-W. Kang and H.-B. Kang, ``Prediction of crime occurrence from multi-modal
  data using deep learning,'' \emph{PloS one}, vol.~12, 2017.

\bibitem{FGSM}
C.~Szegedy, W.~Zaremba, I.~Sutskever, J.~Bruna, D.~Erhan, I.~Goodfellow, and
  R.~Fergus, ``Intriguing properties of neural networks,'' in \emph{ICLR},
  2014.

\bibitem{PGD}
A.~Madry, A.~Makelov, L.~Schmidt, D.~Tsipras, and A.~Vladu, ``Towards deep
  learning models using resistant to adversarial attacks,'' in \emph{ICLR},
  2018.

\bibitem{C-W-attack}
N.~Carlini and D.~Wagner, ``Towards evaluating the robustness of neural
  networks,'' in \emph{IEEE Symposium on Security and Privacy (SP)}, 2017, pp.
  39--57.

\bibitem{Substitute}
N.~Papernot, P.~McDaniel, I.~Goodfellow, S.~Jha, Z.~B. Celik, and A.~Swami,
  ``Practical black-box attacks against machine learning,'' in \emph{ACM on
  Asia Conference on Computer and Communication Security}, 2017, pp. 506--519.

\bibitem{ZOO}
P.-Y. Chen, H.~Zhang, Y.~Sharma, J.-F. Yi, and C.-J. Hsieh, ``Zoo: Zeroth order
  optimization based black-box attacks to deep neural networks without training
  substitute models,'' in \emph{10th ACM Workshop on Artificial Intelligence},
  2017, pp. 15--26.

\bibitem{Orthogonal-bbx}
M.~A. A.~K. Jalwana, N.~Akhtar, M.~Bennamoun, and A.~S. Mian, ``Orthogonal deep
  models as defense against black-box attacks,'' \emph{IEEE Access}, vol.~8,
  pp. 119\,744--119\,757, 2020.

\bibitem{Space-TAE}
F.~Tramèr, N.~Papernot, I.~J. Goodfellow, D.~Boneh, and P.~D. McDaniel, ``The
  space of transferable adversarial examples,'' in \emph{arXiv:1704.03453},
  2017.

\bibitem{Attack-Fool}
N.~Akhtar, M.~A. A.~K. Jalwana, M.~Bennamoun, and A.~S. Mian, ``Attack to fool
  and explain deep networks,'' \emph{IEEE Transactions on Pattern Analysis and
  Machine Intelligence. DOI: 10.1109/TPAMI.2021.3083769}, pp.~--, 2021.

\bibitem{Adversarial-features}
A.~Ilyas, S.~Santurkar, D.~Tsipras, L.~Engstrom, B.~Tran, and A.~Madry,
  ``Adversarial examples are not bugs, they are features,'' in \emph{NeurIPS},
  2019, pp. 125--136.

\bibitem{Feature-scatter}
H.~Zhang and J.~Wang, ``Defense against adversarial attacks using feature
  scattering-based adversarial training,'' in \emph{arXiv:1907.10764}, 2019.

\bibitem{AT-free}
A.~Shafahi, M.~Najibi, A.~Ghiasi, Z.~Xu, J.~Dickerson, C.~Studer, L.~S. Davis,
  G.~Taylor, and T.~Goldstein, ``Adversarial training for free,'' in
  \emph{arXiv:1904.12843}, 2019.

\bibitem{Logit-pairing}
H.~Kannan, A.~Kurakin, and I.~Goodfellow, ``Adversarial logit pairing,'' in
  \emph{arXiv:1803.06373}, 2018.

\bibitem{Ensemble-AT}
F.~TramÃ¨r, A.~Kurakin, N.~Papernot, I.~Goodfellow, D.~Boneh, and
  P.~McDaniel, ``Ensemble adversarial training: Attacks and defenses,'' in
  \emph{ICLR}, 2018.

\bibitem{PNI}
Z.~He, A.~S. Rakin, and D.~Fan, ``Parametric noise injection: trainable
  randomness to improve deep neural network robustness against adversarial
  attack,'' in \emph{CVPR}, 2020, pp. 588--597.

\bibitem{Regularization-robustness}
A.~Bietti, G.~Mialon, and J.~Mairal, ``On regularization and robustness of deep
  neural networks,'' in \emph{arXiv:1810.00363}, 2018.

\bibitem{Input-Grad-R}
A.~S. Ross and F.~Doshi-Velez, ``Improving the adversarial robustness and
  interpretability of deep neural networks by regularizing input gradients,''
  in \emph{AAAI}, 2018.

\bibitem{IntegGrad}
M.~Sundararajan, A.~Taly, and Q.~Yan, ``Gradients of counterfactuals,'' in
  \emph{arXiv:1611.02639}, 2016.

\bibitem{SmoothGrad}
D.~Smilkov, N.~Thorat, B.~Kim, F.~Viegas, and M.~Wattenberg, ``Smoothgrad:
  removing noise by adding noise,'' in \emph{ICML Workshop on Visualization for
  Deep Learning}, 2017.

\bibitem{Robustness-odds}
D.~Tsipras, S.~Santurkar, L.~Engstrom, A.~Turner, and A.~Madry, ``Robustness
  may be at odds with accuracy,'' in \emph{arXiv:1805.12152}, 2018.

\bibitem{Robust-saliency}
C.~Etmann, S.~Lunz, P.~Maass, and C.-B. Schonlieb, ``On the connection between
  adversarial robustness and saliency map interpretability,'' in
  \emph{arXiv:1905.04172}, 2019.

\bibitem{TRADES}
H.~Zhang, Y.~Yu, J.~Jiao, E.~P. Xing, L.~E. Ghaoui, and M.~I. Jordan,
  ``Theoretically principled trade-off between robustness and accuracy,'' in
  \emph{arXiv:1901.08573}, 2019.

\bibitem{Adv-local-linearization}
C.~Qin, J.~Martens, S.~Gowal, D.~Krishnan, A.~Fawzi, S.~De, R.~Stanforth, and
  P.~Kohli, ``Adversarial robustness through local linearization,'' in
  \emph{arXiv:1907.02610}, 2019.

\bibitem{Biological-adversarial}
A.~Nayebi and S.~Ganguli, ``Biologically inspired protection of deep networks
  from adversarial attacks,'' in \emph{arXiv:1703.09202}, 2017.

\bibitem{Input-gradients-robustness}
A.~S. Ross and F.~Doshi-Velez, ``Improving the adversarial robustness and
  interpretability of deep neural networks by regularizing their input
  gradients,'' in \emph{AAAI}, 2018.

\bibitem{Dropconnect}
L.~Wan, M.~Zeiler, S.~Zhang, Y.~LeCun, and R.~Fergus, ``Regularization of
  neural networks using dropconnect,'' in \emph{ICML}, 2013, pp. 1058--1066.

\bibitem{Dropout}
N.~Srivastava, G.~Hinton, A.~Krizhevsky, I.~Sutskever, and R.~Salakhutdinov,
  ``Dropout: a simple way to prevent neural networks from overfitting,''
  \emph{Journal of Machine Learning Research}, vol.~15, pp. 1929--1958, 2014.

\bibitem{Robust-jacobian}
J.~Hoffman, D.~A. Roberts, and S.~Yaida, ``Robust learning with jacobian
  regularization,'' in \emph{arXiv:1908.02729}, 2019.

\bibitem{Adv-examples-physical}
A.~Kurakin, I.~Goodfellow, and S.~Bengio, ``Adversarial examples in the
  physical world,'' in \emph{arXiv:1607.02533}, 2016, pp.~--.

\bibitem{JSMA}
N.~Papernot, P.~McDaniel, S.~Jha, M.~Fredrikson, Z.~B. Celik, and A.~Swami,
  ``The limitations of deep learning in adversarial settings,'' in \emph{IEEE
  European Symposium on Security and Privacy}, 2016, pp. 372--387.

\bibitem{Feature-squeezing}
W.~Xu, D.~Evans, and Y.~Qi, ``Feature squeezing: detecting adversarial examples
  in deep neural networks,'' in \emph{arXiv:1704.01155}, 2017.

\bibitem{Cleverhans}
N.~Papernot, I.~Goodfellow, R.~Sheatsley, R.~Feinman, and P.~McDaniel,
  ``Cleverhans v1.0.0: an adversarial machine learning library,'' in
  \emph{arXiv:1610.00768}, 2016, pp.~--.

\bibitem{Distillation-Defense}
N.~Papernot, P.~McDaniel, X.~Wu, S.~Jha, and A.~Swami, ``Distillation as a
  defense to adversarial perturbations against deep neural networks,'' in
  \emph{IEEE Symposium on Security and privacy}, 2016, pp. 582--597.

\bibitem{Lya-Theory}
A.~Rahnama, A.~T. Nguyen, and E.~Raff, ``Robust design of deep neural networks
  against adversarial attacks based on lyapunov theory,'' in \emph{CVPR}, 2020,
  pp. 8178--8187.

\bibitem{Baehrens-JMLR-2010}
D.~Baehrens, T.~Schroeter, S.~Harmeling, M.~Kawanabe, K.~Hansen, and K.-R.
  Mueller, ``How to explain individual classification decisions,''
  \emph{Journal of machine learning research}, vol.~11, pp. 1803--1831, 2010.

\bibitem{Double-BPP}
H.~Drucker and Y.~L. Cun, ``Improving generalization performance using double
  back-propagation,'' \emph{IEEE Transactions on Neural Networks}, vol.~3, pp.
  991--997, 1992.

\bibitem{Deep-Inside}
K.~Simonyan, A.~Vedaldi, and A.~Zisserman, ``Deep inside convolutional
  networks: Visualising image classification models and saliency maps,'' in
  \emph{arXiv:1312.6034}, 2013, pp.~--.

\bibitem{DIGR}
J.~Xing, T.~Nagata, X.~Zou, E.~Neftci, and J.~L. Krichmar, ``Policy
  distillation with selective input gradient regularization for efficient
  interpretability,'' in \emph{arXiv:2205.08685}, 2022, pp.~--.

\bibitem{Saliency-RL}
M.~Rosynski, F.~Kirchner, and M.~Valdenegro-Toro, ``Are gradient-based saliency
  maps useful in deep reinforcement learning?'' in \emph{NeurIPS 2020
  workshop}, 2020, pp.~--.

\bibitem{JARN}
A.~Chan, Y.~Tay, Y.-S. Ong, and J.~Fu, ``Jacobian adversarially regularized
  networks for robustness,'' in \emph{ICLR}, 2020.

\bibitem{MNIST}
Y.~L. Cun, L.~Bottou, Y.~Bengio, and P.~Haffner, ``Gradient based learning
  applied to document recognition,'' \emph{Proceeding of IEEE}, vol.~86, pp.
  2278--2324, 1998.

\bibitem{CIFAR-10}
A.~Krizhevsky and G.~Hinton, ``Learning multiple layers of features from tiny
  images,'' in \emph{Technique report, Citeseer}, 2009.

\bibitem{ResNet}
K.~He, X.~Zhang, S.~Ren, and J.~Sun, ``Deep residual learning for image
  recognition,'' in \emph{CVPR}, 2016.

\bibitem{Transfer-attack}
Y.~Liu, X.~Chen, C.~Liu, and D.~Song, ``Delving into transferable adversarial
  examples and black-box attacks,'' in \emph{arXiv:1611.02770}, 2016.

\bibitem{DP}
M.~Lecuyer, V.~Atlidakis, R.~Geambasu, D.~Hsu, and S.~Jana, ``Certified
  robustness to adversarial examples with differential privacy,'' in
  \emph{arXiv:1802.03471}, 2018.

\bibitem{RSE}
X.~Liu, M.~Cheng, H.~Zhang, and C.-J. Hsie, ``Towards robust neural networks
  via random self-ensemble,'' in \emph{ECCV}, 2018.

\bibitem{Adv-BNN}
X.~Liu, Y.~Li, C.~Wu, and C.-J. Hsieh, ``Adv-bnn: Improved adversarial defense
  through robust bayesian neural network,'' in \emph{ICLR}, 2019.

\bibitem{Thermometer}
J.~Buckman, A.~Roy, C.~Raffel, and I.~Goodfellow, ``Thermometer coding: one hot
  way to resist adversarial examples,'' in \emph{ICLR}, 2018.

\bibitem{Gradient-Obfuscation}
A.~Athalye, N.~Carlini, and D.~Wagner, ``Obfuscated gradients give a false
  sense of security: circumventing defenses to adversarial examples,'' in
  \emph{arXiv:1802.00420}, 2018.

\end{thebibliography}

\end{document}